\documentclass{article}

\usepackage{arxiv}

\usepackage[utf8]{inputenc} 
\usepackage[T1]{fontenc}    
\usepackage{hyperref}       
\usepackage{url}            
\usepackage{booktabs}       
\usepackage{amsfonts}       
\usepackage{nicefrac}       
\usepackage{microtype}      
\usepackage{lipsum}

\usepackage{hyperref}
\usepackage{cleveref}

\usepackage{graphicx}
 \usepackage{algorithm}
\usepackage{algorithmic}

\usepackage{calrsfs}
\DeclareMathAlphabet{\pazocal}{OMS}{zplm}{m}{n}
\newcommand{\La}{\pazocal{L}}
\newcommand{\Ea}{\pazocal{E}}
\newcommand{\Fa}{\pazocal{F}}
\newcommand{\Aa}{\pazocal{A}}
\newcommand{\Ra}{\pazocal{R}}
\newcommand{\Sa}{\pazocal{S}}
\newcommand{\Ga}{\pazocal{G}}

\newcommand{\Ka}{\pazocal{K}}
\newcommand{\Ta}{\pazocal{T}}

\newcommand{\Ca}{\pazocal{C}}
\newcommand{\Pa}{\pazocal{P}}

\newcommand{\Ba}{\pazocal{B}}

\usepackage{amssymb,amsmath}

\title{An Argumentation-based Approach for Identifying and Dealing with Incompatibilities among\\ Procedural Goals}

\author{
  Mariela Morveli-Espinoza \\
  Graduate Program in Electrical and Computer Engineering (CPGEI),\\
 Federal University of Technology - Paran\'{a} (UTFPR),
 Curitiba - Brazil\\
  \texttt{morveli.espinoza@gmail.com} \\
   \And
 Juan Carlos Nieves \\
  Department of Computing Science of Ume{\aa}  University, 
 Ume{\aa}  - Sweden \\
  \texttt{jcnieves@cs.umu.se} \\
  \AND
   Ayslan Possebom \\
  Graduate Program in Electrical and Computer Engineering (CPGEI),\\
 Federal University of Technology - Paran\'{a} (UTFPR),
 Curitiba - Brazil\\
   \texttt{possebom@gmail.com} \\
   \And
   Josep Puyol-Gruart \\
   Artificial Intelligence Research Institute (IIIA-CSIC), 
 Barcelona - Spain\\
   \texttt{puyol@iiia.csic.es} \\
   \And
   Cesar Augusto Tacla \\
   Graduate Program in Electrical and Computer Engineering (CPGEI),\\
 Federal University of Technology - Paran\'{a} (UTFPR),
 Curitiba - Brazil\\
 \texttt{tacla@utfpr.edu.br} \\
}

\newtheorem{defn}{Definition}
\newtheorem{prop}{Proposition}
\newtheorem{exem}{Example}
\newtheorem{proof}{Proof}
\newtheorem{theorem}{Theorem}

\begin{document}

\maketitle

\begin{abstract}
During the first step of practical reasoning, i.e. deliberation, an intelligent agent generates a set of pursuable goals and then selects which of them he commits to achieve. An intelligent agent may in general generate multiple pursuable goals, which may be incompatible among them. In this paper, we focus on the definition, identification and resolution of these incompatibilities. The suggested approach considers the three forms of incompatibility introduced by Castelfranchi and Paglieri, namely the terminal incompatibility, the instrumental or resources incompatibility and the superfluity. We characterise computationally these forms of incompatibility by means of arguments that represent the plans that allow an agent to achieve his goals. Thus, the incompatibility among goals is defined based on the conflicts among their plans, which are represented by means of attacks in an argumentation framework. We also work on the problem of goals selection; we propose to use abstract argumentation theory to deal with this problem, i.e. by applying argumentation semantics. We use a modified version of the ``cleaner world'' scenario in order to illustrate the performance of our proposal.

\end{abstract}


\section{Introduction}
Practical reasoning means reasoning directed towards actions, i.e. it is the process of figuring out what to do. According to  \hbox{\cite{wooldridge2000reasoning}}, practical reasoning involves two phases: (i) deliberation, which is concerned with deciding what state of affairs an agent wants to achieve, thus, the outputs of deliberation phase are goals the agent intends to pursue, and (ii) means-ends reasoning, which is concerned with deciding how to achieve these states of affairs, thus, the outputs of means-ends reasoning are plans. The first phase is also decomposed in two parts: (i) firstly, the agent generates a set of possible goals, which we call pursuable goals\footnote{Pursuable goals are also known as desires and pursued goals as intentions. In this work, we consider that both are goals at different stages of processing.}, and (ii) secondly, the agent chooses which goals he will be committed to bring about. 

This paper\footnote{This article is an extended version of the extended abstract originally presented at the 16th Conference on Autonomous Agents and MultiAgent Systems, AAMAS'17 \cite{morveli2017dealing}.} focuses on the process of goal selection; i.e. on deciding which consistent set of goals the agent will pursue. We specifically deal with procedural goals\footnote{A goal is called procedural when there is a set of plans for achieving it. This differs from declarative ones, which are a description of the state sought \cite{winikoff2002declarative}. Other authors refer to this type of goal as achievement goal \cite{braubach2004goal}\cite{hubner2006programming}.} and consider that the agents have limited resources, in other words, we work with resource-bounded agents.

Given that an intelligent agent may generate multiple pursuable goals, some incompatibilities among these goals could arise, in the sense that it is not possible to pursue them   simultaneously. Thus, a rational agent should not simultaneously pursue a goal $g$ and a goal $g'$ if $g$ prevents the achievement of $g'$, in other words, if they are inconsistent
\cite{winikoff2002declarative}. Reasons for not pursuing some goals simultaneously are generally related to the fact that plans for reaching such goals may block each other \cite{van2009goals}. 

For a better illustration of the problem, let us present the following scenario. It is based on the well-know ``cleaner world'' scenario, where a set of robots have the task of cleaning the dirt of an environment. The environment is divided into numbered zones referenced by using ordered pairs, which facilitates the communication among the robots. The main goal of all the robots is to have the environment clean. There can be two kinds of dirt, the liquid ones and the solid ones, when it is liquid dirt, the agent mops it and when it is solid dirt, the agent picks it up. All of the robots are equipped with a trash can with limited capacity and have the ability of cleaning both kinds of dirt. However, depending on the distance between the agent and the dirt a robot can or cannot recognize the kind of dirt. In the environment, there is also a workshop, where robots can go to be fixed or to recharge their batteries, and a waste bin, where robots carry the trash they have picked up. During the execution of this task the agents may generate several pursuable goals, which can originate some conflicts or incompatibilities among them. 

According to Castelfranchi and Paglieri \cite{castelfranchi2007role}, three forms of incompatibility could emerge: terminal, instrumental and superfluity. Let's see an example of each kind of incompatibility based on our scenario:

\begin{itemize}
\item \textit{Terminal incompatibility}: Suppose that at a given moment one of the robots -- let us call him $\mathtt{BOB}$ -- detects dirt in slot (3,4); hence, the goal ``cleaning slot (3,4)'' becomes pursuable. On the other hand, $\mathtt{BOB}$ also detects a minor technical defect in his antenna; hence, the goal ``going to the workshop to be fixed'' also becomes pursuable. $\mathtt{BOB}$ cannot pursue both goals at the same time because the plans adopted for each goal lead to an inconsistency, since he needs to be operative to clean slot (3,4) or become non-operative to go to the workshop to be fixed.

\item \textit{Instrumental or resource incompatibility}: It arises because the agents have limited resources. Suppose that $\mathtt{BOB}$ is in slot (1,4) and detects two dirty slots, slot (3,4) and slot (4,1). Therefore, goals ``cleaning slot (3,4)'' and ``cleaning slot (4,1)'' become pursuable; however, he only has battery for executing the plan of one of the goals. Consequently a conflict due to resource battery arises, and $\mathtt{BOB}$ has to choose which slot to clean.

\item \textit{Superfluity}: It occurs when the agent pursues two goals that lead to the same end. Suppose that $\mathtt{BOB}$ is in slot (1,4). On one hand, he detects dirt in slot (5,5), since it is far from its location, he cannot identify the kind of dirt; hence, the goal ``cleaning slot (5,5)'' becomes\break pursuable. On the other hand, another cleaner robot -- let us call him $\mathtt{TOM}$ -- also detects the same dirty slot and he also notes that it is liquid dirt; however, $\mathtt{TOM}$'s battery is low and he is not able to do the task, whereby he sends a message to $\mathtt{BOB}$ to mop slot (5,5); hence, the goal ``mopping slot (5,5)'' becomes pursuable. It is easy to notice that both goals have the same end, which is that slot (5,5) be cleaned, with the difference that ``mopping slot (5,5)'' is a more specific goal. 
\end{itemize}

From these simple examples, one can observe that goals conflict emerge quite easily; hence, if conflicts arise an agent should be able to choose which goals he will pursue, in other words, he should be able to deal with such conflicts or incompatibilities. Besides, BDI-based agent\footnote{BDI is the acronym for Belief-Desire-Intention model \cite{bratman1987intention} \cite{rao1995bdi}.} programming languages should allow agent programmers to implement agents that do not pursue incompatible goals simultaneously, and that can choose from possibly incompatible goals \cite{tinnemeier2007goal}. Therefore, the study of the possible forms of incompatibilities among goals will benefit both the theoretical research and the practical applications.

The notion of conflicts is not something novel. Some researchers have focused on both the detection and the resolution of conflicts, other ones only on the conflicts detection, and others on the resolution of emerging conflicts.  Thangarajah is one of the authors that has worked widely on this problem. In \cite{thangarajah2002representation}, he and his partners propose a general framework for detection and resolution of conflicts; in \cite{thangarajah2003detecting}, they focus on detecting and dealing with a special kind of conflict; and finally, in \cite{thangarajah2002avoiding}, they focus on resolving resource conflicts.  The deliberation strategy for choosing the goals the agent will pursue is the concern of other researches (e.g., \cite{khan2010logical}\cite{pokahr2005goal} \cite{tinnemeier2007goal} \cite{wang2012runtime}\cite{zatelli2016conflicting}). On the other hand, since this is a problem directly related to agent programming languages, \hbox{Zatelli et al. \cite{zatelli2016conflicting}} present a summary of how some languages deal with this problem. Thus, in some of these platforms, the programmer is in charge of specifying which goals must be pursued atomically in order to avoid any kind of conflict with other goals (e.g., Jason \cite{bordini2007programming}, \hbox{2APL \cite{dastani20082apl},} and JIAC \cite{wieczorek1998open}). Other platforms give more flexibility and allow that \hbox{non-atomic} goals be pursued at the same time with atomic goals (e.g., N-2APL \cite{alechina2012programming}, AgentSpeak(RT) \cite{Vikhorev2011}, and ALOO \cite{ricci2012programming}).

Formal argumentation, or just argumentation, is an appropriate approach for reasoning with inconsistent (conflicting) information \cite{dung1995acceptability}. Although argumentation has been usually used for formal reasoning\footnote{Formal reasoning has to do with reasoning about propositional attitudes
such as knowledge and beliefs.}, there are some researches that have applied it for practical reasoning for the generation of desires and plans (e.g., \cite{DBLP:journals/ijar/AmgoudDL11}\cite{amgoud2008constrained}\cite{hulstijn2004combining}\cite{rahwan2006argumentation}). The process of argumentation is based on the construction and the comparison of arguments (considering attacks or conflicts among them) in order to determine the most acceptable of them. The classical form of attack is usually due to the logical inconsistency between the elements that make up two arguments. However, in the resource incompatibility and superfluity the conflict arises due to other reasons as it was presented in the above examples. Thus, we have identified that in the context of practical reasoning, the meaning we give to the arguments and to the attacks can define new forms of conflicts between arguments, which can also be supported by the argumentation inferences. 

Against this background, the aim of this article is to study and formalize the aforementioned three forms of incompatibility, to show how to identify each of them taking into account the plans of the agent and how to deal with them. Our proposal is based on argumentation-based reasoning, since it is a suitable approach for reasoning with inconsistent information \cite{dung1995acceptability}. Thus, the research questions that are addressed in this paper are: (i) Can we identify when a kind of incompatibility arises by using arguments that represent the plans of the agent? if so, how would it be done? and (ii) faced with conflicting plans, how can the set of consistent goals be chosen?.

\begin{figure}[!htb]
	\centering
	\includegraphics[width=0.8\textwidth]{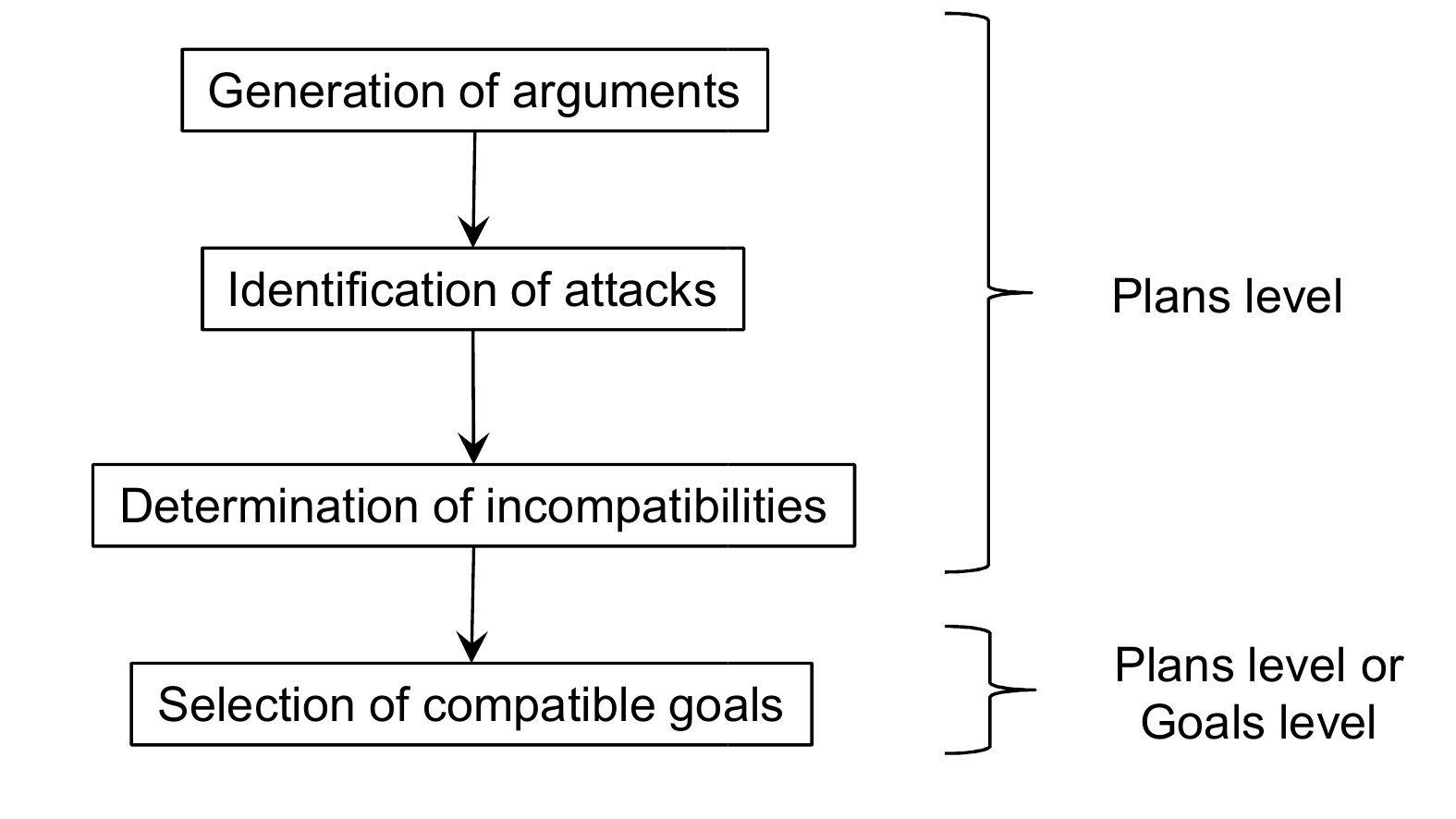} 
	\caption{General view of our argumentation-based approach. We can group the necessary steps in two levels, the plans level and the goals level. The first three steps are exclusively done over a set of plans, which are represented by means of arguments. Once the attacks among arguments are identified, they are used to determine which goals are incompatible. Finally, argumentation semantics are then used in order to find the set of goals that can be pursued without conflicts (final step). }
	\label{workflow}
\end{figure}

In addressing the first question, we start with a set of goals (possible incompatible) such that each goal has a preference value and a set of plans that allow the agent to achieve it. We represent the agent's plans by means of arguments and define the kinds of attacks that determine the incompatibilities. 
Regarding the second question, we use argumentation semantics in order to obtain a set of consistent goals (Figure \ref{workflow} shows the workflow of the proposed argumentation-based approach). In summary, the main theoretical contributions of this paper are: 

\begin{itemize}
\item An arguments-based formalization of the three forms of incompatibility introduced by \cite{castelfranchi2007role}. More specifically, our contribution is on the formalization of the resource incompatibility and the superfluity because these kinds of attacks has not been explored in the state of the art. It is specially in these kinds of attacks that the intended meaning of the arguments in the context of practical reasoning becomes clearer and leads to a novel characterization of attacks because these ones are not based on the evaluation of the inconsistency between logical formulae but they extend the notion of attack.

\item An argumentation-based approach for dealing with the incompatibilities that were identified. We provide two ways for selecting the goals the agent will commit to pursue. The first one is based on the arguments generated by the agent, and the second one is based on the set of pursuable goals. In order to identify the attacks between pursuable goals, we base on the attacks identified between arguments. The proposed approach is an answer to the need of a holistic approach that integrates and deals with more than one type of incompatibility. While it is true that the attacks for determining resource incompatibility and superfluity are new in argumentation theory, these types of conflicts have been already studied from other perspectives (see Related Work in Section \ref{correlatos}); however, different approaches to deal with each of them were employed.

\item A theoretical study of the properties of this formalization; thus, we show that the results of this proposal satisfy the rationality postulates determined in \cite{caminada2007evaluation}, namely consistency and completeness (closure).
\end{itemize}

This work has also a practical contribution since it can be applied to real engineering problems. As it can be seen in the example, this kind of approach can be used in robotic applications (e.g., \cite{davids2002urban}\cite{emmi2014new}\cite{tuna2014autonomous}) in order to endow a robot with a system that allows him to recognize and decide about the goals he should pursue. Another possible application is in the spatial planning problem, which aims to rearrange the spatial environment in order to meet the needs of a \hbox{society \cite{ligtenberg2004design}.} As space is a limited resource, it causes that the planner finds conflicts in the desires and expectations about the spatial environment. These desires and expectation can be modeled as a set of restrictions and conditions, which can be considered as goals (e.g., suitability, dependency, and compatibility)\cite{behzadi2013introducing}. Thus, the planner can be seen as a software agent that has to decide among a set of conflicting goals. Although these conflicting goals are not goals the agent wants to achieve, as in the case of the robot, the agent may use this approach in order to resolve the problem and suggest a possible arrangement of the spatial environment. It could also be applied during a design process, in which inconsistencies among design objectives may arise and this results in design conflicts \cite{canbaz2014preventing}. In this case, agents may represent designers that share knowledge and have conflicting interests. This results in a distributed design system that can be simulated as an Multi-Agent System \cite{chira2005multi}\cite{fan2008development}. This last type of application involves more than one agent; however, since there is a conflict among goals, it can be resolved by applying the proposed approach.

The rest of the paper is organized as follows. Next section introduces basic concepts about the arguments on which this approach is based. In Section \ref{ataques}, we define what kinds of attacks may occur between arguments, which will lead to the identification of each form of incompatibility between goals. In Section \ref{postulados}, we delineate a set of properties related to the attacks. This is important because it will allow us to use the concepts of abstract argumentation in our approach.   
Section \ref{frames} is focused on the definition of argumentation frameworks, one for each kind of incompatibility and a general argumentation framework. In Section \ref{seleccion}, we study how to determine the set of compatible goals by means of argumentation semantics applied to the argumentation frameworks. We present the evaluation of our proposal in terms of fulfilling the postulates of rationality in Section \ref{evalua}. Finally, related work is discussed in Section \ref{correlatos} and the conclusions and future work are presented in Section \ref{conclus}. All the proofs are given in an appendix at the end of the document.

\section{Background}

In this section, we will recall basic concepts related to the abstract argumentation framework (AF) developed by Dung \cite{dung1995acceptability}, including the notion of acceptability and the some semantics. The reader with previous knowledge on abstract argumentation can skip reading this section. This section does not aim to be a tutorial on abstract argumentation.

\begin{defn}\label{def-dung} \textbf{(Argumentation Framework)} An argumentation framework $\Aa\Fa$ is a tuple $\Aa\Fa = \langle \Aa rg, \Ra \rangle$ where $\Aa rg$ is a finite set of arguments and $\Ra$ is a binary relation $\Ra$ $\subseteq \Aa rg \times \Aa rg$ that represents the attack between two arguments of $\Aa rg$, so that $(A,B) \in \Ra$ denotes that the argument $A$ attacks the argument $B$.

\end{defn}

In argumentation theory, an acceptability semantics is a function in charge of returning sets of arguments called extensions which are internally consistent. Next, we introduce the concepts of conflict-freeness, defense, admissibility and the main semantics that we will later analyse in the Section \ref{seleccion} in order to determine which semantics (or family of semantics) is the most adequate for the goals selection process.

\begin{defn} \textbf{(Basic concepts)}\label{basicosarg} Let $\Aa\Fa=\langle \Aa rg, \Ra \rangle$ an argumentation framework and a set $\Ea \subseteq \Aa rg$:

\begin{itemize}
    \item $\Ea$ is \textit{conflict-free} if $\forall A, B \in \Ea$, $(A, B) \notin \Ra$.
\item $\Ea$ \textit{defends} an argument $A$ iff for each argument $B \in \Aa rg$, if $(B, A)$, then there exist an argument $C \in \Ea$ such that $(C, B)$.
\item $\Ea$ is \textit{admissible} iff it is conflict-free and defends all its elements. 
\end{itemize}
\end{defn}

Next, we define preferred semantics. This semantics is based on the notion of admissibility and with the idea of maximizing the accepted arguments. We have chosen this semantics because it can be considered a representative semantics of the family of the admissibility-based semantics.

\begin{defn} \textbf{(Preferred semantics)} Given an argumentation framework $\Aa\Fa=\langle \Aa rg, \Ra \rangle$ and a set $\Ea \subseteq \Aa rg$. $\Ea$ is a preferred extension if it is a maximal (with respect to the set inclusion) admissible subset of $\Aa rg$.

\end{defn}

Unlike preferred semantics, stage semantics are not based on admissibility. The concept of stage semantics has been introduced in \cite{verheij1996two} and further developed in \cite{verheij2003deflog} and, in essence, a stage extension is based on conflict-freeness\footnote{We call of conflict-free semantics to the semantics that is only based on the conflict-freeness concept.}. We have chosen this semantics in order to analyse semantics that are mainly based only on conflict-freeness and the range concept. This semantics was also characterized in terms of 2-valued logical models in \cite{osorio2016range}.

\begin{defn} \textbf{(Stage semantics)} Given an argumentation framework $\Aa\Fa=\langle \Aa rg, \Ra \rangle$ and a set $\Ea \subseteq \Aa rg$, the range of $\Ea$ is defined as $\Ea \cup \Ea^+$, where $\Ea^+ = \{A \;| \; A \in \Aa rg$ and $(B, A) \in \Ra$ and $B \in \Ea\}$. $\Ea$ is a stage extension iff $\Ea$ is a conflict-free set with maximal (with respect to inclusion) range\footnote{More details about this semantics can be found in \cite{verheij1996two}.}. 

\end{defn}
\section{Theoretical Framework}
\label{bloques}

In this section, we present the main mental states of the agent and the representation of his plans by means of arguments, which are called instrumental arguments. 

We start by presenting the propositional logical language that will be used. Let $\La$ be a propositional language used to represent the mental states of the agent, $\vdash$ stands for the inference of classical propositional logic, $\top$ and $\perp$ denote truth and falsum respectively, and $\equiv$ denotes classical equivalence. We use lowercase roman characters to denote atoms and uppercase Greek characters to denote formulae, such that an atomic proposition $b$ is a formula. If $b$ is a formula, then so is $\neg b$. If $b$ and $c$ are formulae, then so are $b \wedge c, b \vee c$, and $b \rightarrow c$. Finally, if $b$ is a formula, then so is $(b)$.

From $\La$, we can distinguish the following finite sets:

\vspace{0.2cm}
- The set $\Ba$, which denotes the beliefs of the agent. \\
\indent - The set $\Ga$, which denotes the goals of the agent.\\
\indent - The set $\Ra\Ea\Sa$, which denotes the resources of the agent.\\
\indent - The set $\Aa$, which denotes the actions of the agent.
\vspace{0.2cm}

\noindent$\Ba$, $\Ga$, $\Ra\Ea\Sa$, and $\Aa$ are subsets of literals\footnote{A literal is either an atomic formula or the negation of an atomic formula. When a literal is an atomic formula, we say that it is a positive literal, and when a literal is the negation of an atomic formula, we say it is a negative literal.} from the language $\La$. It also holds that $\Ba,\Ga,\Ra\Ea\Sa$, and $\Aa$ are pairwise disjoint. 

\begin{exem} Considering the scenario of the cleaner world and example given in the introduction, we next present some possible beliefs, goals, resources, and actions of agent BOB:

\begin{itemize}
    \item \textbf{Beliefs:} there is solid dirt in slot (3,4), the trash can is full, and slot (1,2) is clean.
\item \textbf{Goals:} clean a given slot, recharge battery, and clean the whole environment.
\item \textbf{Resources:} battery, oil, and spare part.
\item \textbf{Actions:} go to the next slot, use the spin mop to clean a given slot, and empty the trash can.
\end{itemize}
\end{exem}

The agent is also equipped with a set of plans that allow him to achieve his goals. In order to analyze the possible conflicts that may arise among goals, we express the agent's plans in terms of arguments, which are called instrumental arguments. The use of instrumental arguments for representing plans is not a novelty. Rahwan and Amgoud \hbox{\cite{rahwan2006argumentation}} define this kind of argument, which is structured like a tree where the nodes are planning rules\footnote{In this work, a plan rule is a building block structure that is used to construct a partial plan, which in turn is a building block that is used to construct an instrumental argument. It is important to differentiate the plan rules used in this work with the plan rules used in classical planning. In this work, the plan rules make up already defined plans, whereas the plan rules used in classical planning support the generation of new plans.} whose components are a set of desires and a set of resources in the premise and a desire in the conclusion. Analogously, we use a set of goals and resources in the premise (goals in the premise can be seen as sub-goals) and a goal in the conclusion. Additionally, we also include a set of beliefs and a set of actions, because if the agent wants to achieve the goal in the conclusion of the rule he needs that some beliefs are true and he also needs to be able to perform some actions. For example, for reaching the goal ``\textit{be fixed}'' it is necessary that a certain spare part is available. Thus, agent $\mathtt{BOB}$ needs that the belief ``\textit{available spare part}'' holds true in order to achieve ``\textit{be fixed}''. Regarding the actions, in order to achieve a goal it is necessary that some actions be performed. For instance, for agent $\mathtt{BOB}$ clean all the environment it is necessary that he moves from one slot to the next. Thus, in this work, a plan rule consists of a finite set of beliefs, a finite set of goals, a finite set of actions, and a finite set of resources in its premise

It is also important to highlight that some of the sets of the premise of a plan rule may be empty. For example, not all of the goals always have sub-goals because if it would happen, it would cause an infinite sequence of calls to sub-plans and a top goal would never be reached.

These plan rules can be the result of an automated planner or can be obtained by gathering information from experts of an application domain. For instance, in \cite{guerrero2018activity,nieves2013reasoning}, the authors propose an approach where fragments of human activities are built from a set of observations (beliefs) of the world, a finite set of actions. We can compare a fragment of an activity with an instrumental argument; hence, a fragment of activity defines a context of a given goal such as a plan rule.

Regarding the resources in the premise of a plan rule, the necessary resource and its necessary amount varies for each plan rule. Thus, let $\Ra\Ea\Sa_{qua}$ be an infinite set of ground atoms that denote a given\break resource along with a given quantity, which is expressed numerically. Then, we have that $\Ra\Ea\Sa_{qua}=\{res\_q(name,value) | res(name) \in \Ra\Ea\Sa, value \in  \mathbb{N}\}$. For example, assume that $\Ra\Ea\Sa=\{res(bat)\}$, where $bat$ is the name that denotes the resource battery. We may have $\Ra\Ea\Sa_{qua}' =\{res\_q(bat,10), res\_q(bat,50)\}$ such that $\Ra\Ea\Sa_{qua}' \subset \Ra\Ea\Sa_{qua}$ and the ground atoms $res\_q(bat,10)$ and $res\_q(bat,50)$ denote that 10 units of battery and 50 units of battery are necessary, respectively. 

Notice that we use the suffix $res\_q$ for denoting resources in $\Ra\Ea\Sa_{qua}$ and the suffix $res$ for denoting resources in $\Ra\Ea\Sa$.

\begin{defn} \textbf{(Plan rule)} A plan rule $pr$ is an expression of the form $b_1  \wedge ... \wedge b_n \wedge g_1 \wedge ... \wedge g_m \wedge a_1  \wedge ... \wedge a_l \wedge res\_q(name,value)_1 \wedge ... \wedge res\_q(name,value)_v\rightarrow g$, where $b_i \in \Ba$ (for all $1 \leq i \leq n$), $g, g_j \in \Ga$, $g \neq  g_j$, (for all $1 \leq j \leq m$), $a_k \in \Aa$ (for all $1 \leq k \leq l$), and $res\_q(name,value)_u \in \Ra\Ea\Sa_{qua}$ (for all $1 \leq u \leq v$). 
\end{defn}
It expresses that $g$ can be achieved\footnote{Achievement goals represent a desired state that an agent wants to reach \cite{dastani2011rich}.} if beliefs $b_1  \wedge ... \wedge b_n$ are true, sub-goals $g_1 \wedge ... \wedge g_m$ can be achieved, actions $a_1  \wedge ... \wedge a_l$ can be performed, and resources $res\_q(name,value)_1 \wedge ... \wedge res\_q(name,value)_v$ are available. It is important to state that the number of elements in the body of a plan rule is finite. Finally, let $\Pa\Ra$ be the base containing the set of plan rules. 

Since there may be goals in the premise of a plan rule, this means that top goals (i.e. goals in the conclusion) may be decomposed into sub-goals, which in turn can be decomposed into sub-sub-goals. Each of these sub-goals has also a plan rule associated to it, which means that there is (at least) a plan rule for each goal.

\begin{exem}\label{ejemprs} Considering the scenario presented in the introduction section, let us introduce some examples of plan rules. Suppose that the environment is a square of 5 $\times$ 5. For the environment to be completely clean, all the slots have to be clean. 

Let us present the beliefs, actions, goals, and resources that make part of the premises of the plan rules. The beliefs the agent should hold are: $has(spare\_part), \neg full\_trashcan,  solid\_dirt(1,1), ..., solid\_dirt(5,5),\break liquid\_dirt(1,1), ..., liquid\_dirt(5,5), unknown\_dirt(1,1), unknown\_dirt(1,2), ..., unknown\_dirt(5,5),$ and $be(operative)$. The actions the agent should perform are: $go(1,1), ..., go(5,5), use(vacuum),use(spinmop)$, and $go(workshop)$. The goals the agent should achieve are: $clean(1,1),..., clean(5,5), clean, mop(1,1), ..., mop(5,5),\break pickup(1,1), ..., pickup(5,5),  be(in\_workshop)$, and $be(fixed)$. We use goal $clean$ to refer to the environment as a whole and we use goals $clean(1,1), ..., clean(5,5)$ to refer to each slot of the environment. Thus, to achieve the goal $clean$, all the dirty slots have to be cleaned. Lastly, the resources that have to be available are $res\_q(bat,40), res\_q(bat,80), res\_q(bat,70)$, and $res\_q(bat,30)$. We assume that the agent needs 10 units of battery to go from a slot to the next, 20 units of battery to use the vacuum, and 10 units of battery to use the mop.\\

\noindent(1) $clean(1,1) \wedge clean(1,2) \wedge ... \wedge clean(5,5) \rightarrow clean$\\
(2) $mop(1,1) \rightarrow clean(1,1), ...,mop(5,5) \rightarrow clean(5,5)$ \\
(3) $pickup(1,1) \rightarrow clean(1,1), ..., pickup(5,5) \rightarrow clean(5,5)$\\
(4) $\neg solid\_dirt(1,1)  \wedge \neg liquid\_dirt(1,1) \wedge \neg unknown\_dirt(1,1) \rightarrow clean(1,1),...,$\\
\hspace*{0.65cm}$\neg solid\_dirt(5,5)  \wedge \neg liquid\_dirt(5,5) \wedge \neg unknown\_dirt(5,5) \rightarrow clean(5,5)$\\
(5) $ has(spare\_part) \wedge be(in\_workshop) \rightarrow be(fixed)$\\
(6) $be(operative) \wedge \neg full\_trashcan \wedge solid\_dirt(3,4) \wedge go(3,4) \wedge use(vacuum)$ \\
\hspace*{0.65cm}$\wedge at(1,4)\wedge res\_q(bat,40) \rightarrow pickup(3,4), $\\
 \hspace*{0.65cm}- $be(operative) \wedge \neg full\_trashcan \wedge solid\_dirt(4,1)   \wedge go(4,1) \wedge use(vacuum)$\\
\hspace*{0.65cm}$ \wedge at(1,4) \wedge res\_q(bat,80) \rightarrow pickup(4,1)$\\
 \hspace*{0.65cm}- $be(operative) \wedge \neg full\_trashcan \wedge solid\_dirt(5,5)  \wedge go(5,5) \wedge use(vacuum)$\\
\hspace*{0.65cm}$ \wedge at(1,4) \wedge res\_q(bat,70) \rightarrow pickup(5,5)$\\
(7) $be(operative) \wedge\neg full\_trashcan \wedge liquid\_dirt(5,5) \wedge go(5,5) \wedge use(spinmop)$\\
\hspace*{0.65cm}$ \wedge at(1,4) \wedge res\_q(bat,60)\rightarrow mop(5,5)$\\
(8) $\neg be(operative) \wedge res\_q(bat,30) \wedge go(workshop) \rightarrow be(in\_workshop)$

  \end{exem}
  
The first plan rule is associated to the general goal of the agent; it is to have cleaned a given environment. The next three plan rules are related to the type of dirt the robot has to clean or the absence of dirt. Plan rule number five expresses what the robot needs in order to be fixed. The next two plan rules expresses what the robot needs in order to achieve goals mopping and picking up dirt of a given slot. Plan rule eight expresses what is necessary for the robot get the workshop, for the sake of simplicity, $go(workshop)$ summarizes a set of actions (e.g. turn to the left, walk one slot, etc.).

We can now define the architecture of an intelligent agent, which is an instantiation of BDI model.

\begin{defn}\label{def-agente} \textbf{(Agent)} An intelligent agent is a tuple $\langle \Ka\Ba,\Pa\Ra, \Ga_p , \mathtt{PREF}, \Ra\Ea\Sa_{sum}\rangle$ where:

\begin{itemize}
\item $\Ka\Ba= \Ba \cup \Aa \cup \Ra\Ea\Sa_{qua}$ is the knowledge base of the agent,
\item $\Pa\Ra$ is the set of plan rules,
\item $\Ga_p \subseteq \Ga$ is the set of pursuable goals\footnote{We do not consider any temporal information about the order in which the goals should be pursued or the time the goals have to achieved.},
\item $\mathtt{PREF:} \Ga \rightarrow [0,1]$ is a function that returns the preference value of a given goal such that 0 stands for the minimum preference value and 1 for the maximum one,
\item $\Ra\Ea\Sa_{sum}  \subset \Ra\Ea\Sa_{qua}$ is a resource summary, which contains the information about the available amount of every resource of the agent. We assume that $\Ra\Ea\Sa_{sum}$ is normalised so that each resource appears exactly once and that all the resources represented in $\Ra\Ea\Sa$ have their corresponding available amount in $\Ra\Ea\Sa_{sum}$. Let $\rho: \Ra \Ea \Sa \rightarrow \mathbb{N}$ a function that returns the currently available amount of a given resource; thus, $\rho(res(name))$ denotes the availability of resource $res(name)$.
\end{itemize}

\end{defn}

Based on his knowledge base $\Ka\Ba$ and the set plan rules $\Pa\Ra$, the agent can build partial plans, which are the building blocks of instrumental arguments, which in turn represent complete plans. The idea is that each element of a complete plan, namely a belief, an action, a resource, or a goal, is represented by a partial plan, which can be seen as a standard argument, whose claim may be a belief, an action, a resource, or a goal.

\begin{defn} \textbf{(Partial plan)} A partial plan $pp$ is a pair of the form $[H, \psi]$ where:

\begin{itemize}

\item $\psi \in \Aa$ and $H=\emptyset$, or
\item $\psi \in \Ba$ and $H=\emptyset$, or
\item $\psi \in \Ra\Ea\Sa_{qua}$ and $H=\emptyset$, or
\item $\psi \in \Ga$ and $H=\{b_1, ... , b_n, g_1, ... , g_m, a_1, ..., a_l, res\_q(name,value)_1, ..., res\_q(name,value)_v\}$ such that $\exists (b_1  \wedge ... \wedge b_n \wedge g_1 \wedge ... \wedge g_m \wedge a_1  \wedge ... \wedge a_l \wedge res\_q(name,value)_1 \wedge ... \wedge res\_q(name,value)_v\rightarrow \psi) \in \Pa\Ra$ 

\end{itemize}
\end{defn}

A partial plan $[H,\psi]$ is called elementary when $H=\emptyset$. Let us call $H$ the support of the partial plan and $\psi$ its conclusion. As for notation, $\mathtt{HEAD}(pp)=\psi$ and $\mathtt{BODY}(pp)=H$ denote the conclusion and support of the partial plan $[H, \psi]$, respectively.  

\begin{exem} Let us express as partial plans some of the plan rules introduced by Example \ref{ejemprs}:

$pp=[\{\neg be(operative), has(spare\_part), be(in\_workshop)\}, be(fixed)]$\\
\indent$pp'=[\{\}, go(workshop)]$\\
\indent$pp''=[\{\}, solid\_dirt(3,4)]$

\vspace{0.1cm}
\noindent Partial plan $pp$ has in its conclusion a goal, and the partial plans $pp'$ and $pp''$ have an action and a belief, respectively.

\end{exem}

Based on partial plans, we can now define instrumental arguments, which correspond to complete plans.

\begin{defn} \textbf{(Instrumental argument, or complete plan)} An instrumental argument is a pair $[\Ta, g ]$ such that $g \in \Ga$, and $\Ta$ is a finite tree such that:
\begin{itemize}

\item The root of the tree is a partial plan $[H,g]$,

\item A node $[\{b_1  \wedge ... \wedge b_n \wedge g_1 \wedge ... \wedge g_m \wedge a_1  \wedge ... \wedge a_l\wedge res\_q(name,value)_1 \wedge ... \wedge res\_q(name,value)_v\}, h']$ has exactly $(n+m+l+v)$ children $[\emptyset,b_1], ..., [\emptyset,b_n]$, $[H_1',g_1], ..., [H_m',g_m]$, $[\emptyset,a_1], ..., [\emptyset,a_l]$, $[\emptyset,res\_q(name,value)_1], ..., [\emptyset,res\_q(name,value)_v]$ where each $[\emptyset,b_i]$ ($1 \leq i \leq n$), $[H_j',g_j]$ ($1 \leq j \leq m$), $[\emptyset,a_k]$ ($1 \leq k \leq l$), $[\emptyset,res\_q(name,value)_u]$ ($1 \leq u \leq v$) is a partial plan,

\item The leaves of the tree are elementary partial plans.

\end{itemize}
Let $\Aa rg$ be the set of all instrumental arguments that can be built from the knowledge base of the agent. We assume that each goal in $\Ga$ has at least one instrumental argument. We will use function $\mathtt{SUPPORT}(A)$ to return the set of partial plans of $\Ta$ and $\mathtt{CLAIM}(A)$ to return the claim $g$ of an instrumental argument $A$. We also use the following function to return the set of arguments whose claim is a given goal: $\mathtt{ARG}(g)= \bigcup \{A | \mathtt{CLAIM}(A)=g$ and $A \in \Aa rg\}$.

\end{defn}

As we said before, instrumental arguments have been already employed for representing plans. We have defined an intrumental argument in the same way as it is defined in \cite{rahwan2006argumentation}, i.e., as a tree of partial plans. However, there are two differences between our definition and the definition given in \cite{rahwan2006argumentation}, which are mainly related to the elements of the partial plans. Thus, in \cite{rahwan2006argumentation}, the conclusions of elementary partial plans are only beliefs whereas according to our definition, the elementary partial plans may have as conclusions beliefs, actions, or resources. The other difference is related to the elements of the non-elementary partial plans. Since a non-elementary partial plan is built from a plan rule, it may have in its premise actions and resources, besides beliefs and goals, which are the elements considered in \cite{rahwan2006argumentation}.

\begin{exem} \label{ejm-arg1} Figure \ref{argumentos1} shows two instrumental arguments. Argument $C$ represents a complete plan for goal $clean(5,5)$ and argument $D$ represents a complete plan for goal $mop(5,5)$. Notice that argument $D$ is a sub-argument of $C$.

\begin{figure}[!htb]
	\centering
	\includegraphics[width=1\textwidth]{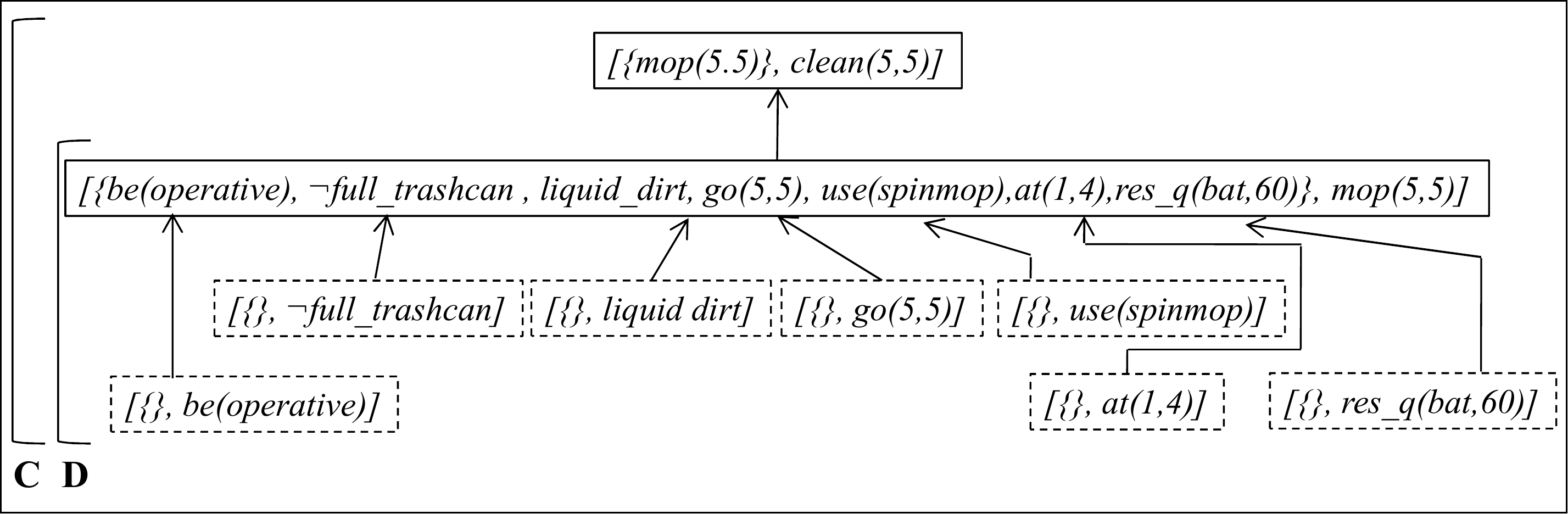} 
	\caption{Instrumental arguments $C$ and $D$ for Example \ref{ejm-arg1}. Dashed-border squares represent the leaves of the three.}
	\label{argumentos1}
\end{figure}
\end{exem}

\section{Attacks between Arguments}
\label{ataques}

In this section, we focus on the identification of attacks among instrumental arguments, which will lead to the identification of each form of incompatibility among goals. The kind of attack depends on the form of incompatibility. We have identified one type of attack for each form of incompatibility. These conflicts between arguments are defined over $\Aa rg$ and are captured by the binary relation $\Ra_x \subseteq \Aa rg \times \Aa rg$ (for $x \in \{t,r,s\}$) where each sub-index denotes the form of incompatibility. Thus, $t$ denotes the attack for terminal incompatibility, $r$ the attack for resource incompatibility, and $s$ the attack for superfluity. We denote with $(A, B)$ the attack relation between arguments $A$ and $B$. In other words, if $(A,B) \in \Ra_x$, it means that argument $A$ attacks argument $B$.

\subsection{Rebuttal between partial-plans}

We can define the terminal incompatibility in terms of attacks among instrumental arguments. In this attack, the beliefs, the goals, and the actions of each partial plan of an argument are taken into account. Thus, an instrumental argument $A$ attacks another instrumental argument $B$ when the conclusion of a partial plan of $A$ is the negation of the conclusion of a partial plan of $B$ and both arguments correspond to plans that allow to achieve different goals. It is important to make two remarks: (i) the nature of the two conclusions has to be the same, i.e. both conclusions have to be or beliefs, or actions, or goals, and (ii) resources are not taken into account in this conflict.  Formally:

\begin{defn} \textbf{(Partial-plans rebuttal - $\Ra_t$)} Let $A,B \in \Aa rg$ be two arguments, $[H, \psi] \in \mathtt{SUPPORT}(A)$ and $[H', \psi'] \in \mathtt{SUPPORT}(B)$ be two partial plans. We say that $(A,B) \in \Ra_t$ occurs when:

\begin{itemize}
\item $\mathtt{CLAIM}(A) \neq \mathtt{CLAIM}(B)$,
\item $\psi = \neg \psi'$ such that $\psi, \psi' \in \Ba$ or $\psi, \psi' \in \Aa$, or $\psi, \psi' \in \Ga$.
\end{itemize}

\end{defn}

We can observe that $\Ra_t$ is symmetric.

\begin{prop}\label{propsimt} If $(A,B) \in \Ra_t$, then $(B, A) \in \Ra_t$.

\end{prop}

\begin{figure}[!htb]
	\centering
	\includegraphics[width=1\textwidth]{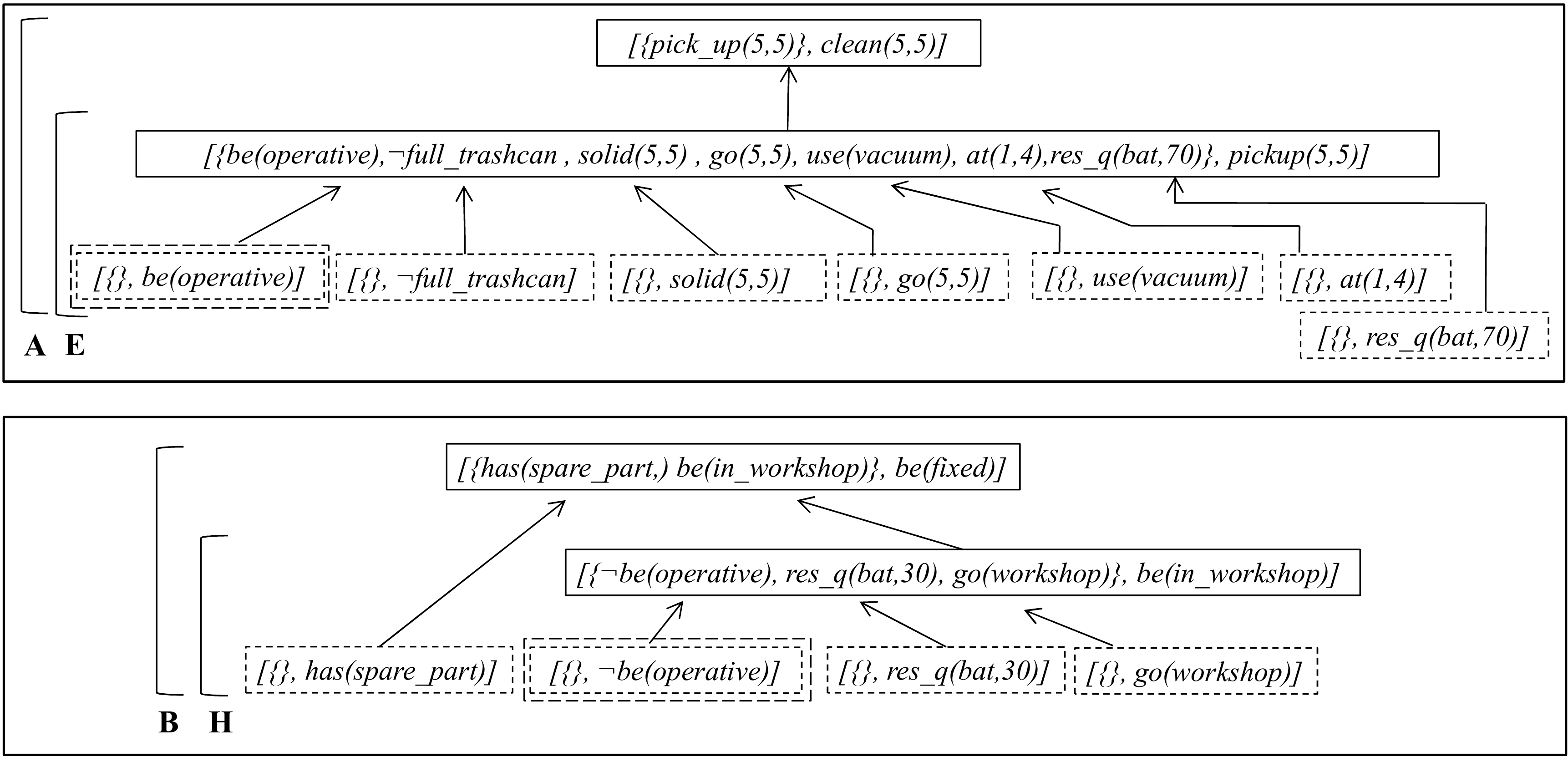} 
	\caption{Arguments $A, E, H$, and $B$ for Example \ref{excleanfix}. There is a partial-plans rebuttal between arguments $A$ and $B$, $A$, and $H$, $E$ and $B$, and $E$ and $H$. The double-border squares highlight the conflicting partial plans. Dashed-border squares represent the leaves of the three.}
	\label{argumentos}
\end{figure}

\begin{exem} \label{excleanfix}Let $\Ga_p=\{clean(5,5),be(fixed)\}$ be two pursuable goals of robot $\mathtt{BOB}$. Figure \ref{argumentos} shows argument $A$ for goal $clean(5,5)$ and argument $B$ for goal $be(fixed)$. Consider also argument $C$, of Figure \ref{argumentos1}, for goal $clean(5,5)$. We can observe three sub-arguments: $E$, whose claim is goal $pickup(5,5)$, is the sub-argument of $A$, $H$, whose claim is $be(in\_workshop)$, is the sub-argument of $B$, and $D$, whose claim is goal $mop(5,5)$, is the sub-argument of $C$. From these arguments, the partial-plans rebuttals that can be identified are: $\Ra_t= \{(A,B), (B,A), (E,B), (B,E), (E,H),(H,E),(A,H),\break(H,A), (C,B),(B,C), (D,B),(B,D), (D,H), (H,D), (C,H),(H,C)\} $.

\end{exem}\label{ejmrebuttal}

\subsection{Attack for identifying the resources incompatibility}

Two arguments are incompatible due to resources because the agent has no enough resources for performing the plans represented by both arguments. Thus, the attack due to resources between two arguments has to reflect this fact. In order to deal with resource conflict, we first define a resource consumption inference that works exclusively for reasoning about resources. This inference considers the availability of a given resource and the amount of it that is necessary. Recall that function $\rho$ --introduced in Definition \ref{def-agente}-- returns the available amount of a given resource; however, the necessary amount has to be obtained from the two arguments whose resource incompatibility is being evaluated. The following steps are carried out in order to obtain this value.

\begin{enumerate}
\item First of all, we put together all the same necessary resources of the two arguments in a formula (let us call it $\Phi_{res(name)}$). This means that there is a different $\Phi_{res(name)}$ for each different resource that both arguments need. Thus, the formula $\Phi_{res(name)}$ is a conjunction of atoms that represent a resource and the necessary amount of it. Such atoms are part of one or more plan rules that make up the two arguments. Hence, we have that $\Phi_{res(name)}= \bigwedge res\_q(name,value)$ where $res\_q(name,value) \in \Ra\Ea\Sa_{qua}$. For example, argument $A$ needs 70 units of battery and argument $B$ needs 30 units of battery; hence, $\Phi_{res(bat)}=res\_q(bat,70) \wedge res\_q(bat,30)$. 

\item The second step is related to the signature of $\Phi_{res(name)}$. Let us denote the signature of $\Phi_{res(name)}$ by $\La_{\Phi_{res(name)}}$. Continuing with the example, $\La_{\Phi_{res(bat)}}=\{res\_q(bat,70), res\_q(bat,30)\}$.

\item Finally, we can sum up the necessary amount of a given resource: $\pi(\La_{\Phi_{res(name)}})= \sum_{[res\_q(name,value) \in \La_{\Phi_{res(name)}}]} value$. Finalizing the example, we have that $\pi(\La_{\Phi_{res(bat)}})=100$. 

\end{enumerate}

Once we have the available amount and the necessary amount of a given resource, we can define the resource-consumption inference. This type of inference resembles other consumption inferences introduced by other consumption and production resources logics like \cite{bulling2010decidability}.

\begin{defn} \textbf{(Resource-consumption inference - $\vdash_r$)} Let $\Ra\Ea\Sa_{sum}$ be the set of available resources of the agent and $\Phi_{res(name)}$ be a conjunction of atoms such that $\La_{\Phi_{res(name)}} \subset \Ra\Ea\Sa_{qua}$. $\Ra\Ea\Sa_{sum}$ satisfies a formula $\Phi_{res(name)}$ (denoted by $\Ra\Ea\Sa_{sum} \vdash_r \Phi_{res(name)}$)  when $\rho(res(name)) \geq \pi(\La_{\Phi_{res(name)}})$.

\end{defn}

The following notation will be used for defining the resource attack. $\mathtt{REC}(A)$ denotes the set of  resources necessary for an argument $A$:

\begin{center}
$\mathtt{REC}(A)=\bigcup_{pp \in \mathtt{SUPPORT}(A)} \mathtt{BODY}(pp) \cap \Ra\Ea\Sa_{qua}$
\end{center}

\begin{defn}\textbf{(Resource attack - $\Ra_r$)} Let $A, B \in \Aa rg$ be two instrumental arguments, $\mathtt{REC}(A)$ be the set of resources necessary for argument $A$, and $\mathtt{REC}(B)$ be the set of resources necessary for argument $B$. We say that $(A,B) \in \Ra_r$ occurs when:

\begin{itemize}
\item $\exists res(name) \in \Ra\Ea\Sa$ such that $\exists res\_q(name,value) \in \mathtt{REC}(A)$ and $\exists res\_q(name,value)' \in \mathtt{REC}(B)$,
\item $\Phi_{res(name)} = \bigwedge_{[res\_q(name,value) \in \mathtt{REC}(A), res\_q(name,value)' \in \mathtt{REC}(B)]} res\_q(name,value)\break \wedge res\_q(name,value)'$,
\item $\Ra\Ea\Sa_{sum} \nvdash_r \;\Phi_{res(name)}$, this means that $\Phi_{res(name)}$ is resource-inconsistent.
\end{itemize}

\end{defn}

We can see that $\Ra_r$ is symmetric.

\begin{prop}\label{propsimr} If $(A, B) \in \Ra_r$, then $(B, A) \in \Ra_r$.

\end{prop}

\begin{exem} \label{ejm-recursos} Let us recall that $\Ga_p=\{clean(5,5), be(fixed)\}$ are the two pursuable goals of agent $\mathtt{BOB}$. Figures \ref{argumentos1} and \ref{argumentos} show arguments $C, A$ and $B$. Notice that argument $C$ needs 60 units of battery ($res\_q(bat,60)$), argument $A$ needs 70 units of battery ($res\_q(bat,70)$), and argument $B$ needs 30 units of battery ($res\_q(bat,30)$). Recall that $\Ra\Ea\Sa_{sum}=\{res\_q(bat,90), res\_q(oil,50),\break res\_q(fuel,20)\}$.

Notice that for achieving goal $clean(5,5)$, there are two arguments with different needs of battery namely $C$ and $A$, and for achieving goal $be(fixed)$ there is only one argument namely $B$. Considering arguments $C$ and $B$, we have $\Phi_{res(bat)} = res\_q(bat,60) \wedge res\_q(bat,30)$, in this case we can say that $\Ra\Ea\Sa_{sum} \vdash_r \Phi_{res(bat)}$ because the agent has enough resources for performing both plans. Otherwise, considering arguments $A$ and $B$, we have $\Phi'_{res(bat)} = res\_q(bat,70) \wedge res\_q(bat,30)$, in this case we can say that $\Ra\Ea\Sa_{sum} \nvdash_r \Phi'_{res(bat)}$ because there is no enough energy for performing both plans. Note that $\Phi'_{res(bat)}$ is the same for arguments $E$ and $B$; therefore, we can say that exists resource attack between $A$ and $B$, and also between $E$ and $B$. The same reasoning applies to arguments $A$ and $H$ and $E$ and $H$. Thus, we have the following attack relations: $\Ra_r= \{(A,B), (B,A), (E,B), (B,E), (A,H),(H,A),\break (E,H),(H,E)\}$.

\end{exem}

\subsection{Superfluous conflict}

Superfluity emerges when two plans lead the agent to the same end, in other words, when two arguments have the same claim. Unlike other contexts, in practical reasoning the fact that two arguments support the same claim is considered unnecessary, or even worse, a waste of time or resources, because it means that the agent performs two plans when only one is necessary for achieving a given goal. Superfluity can be defined in terms of non-elementary partial plans, i.e. in terms of partial plans whose conclusions are goals. Before presenting the definition of superfluous attack, we will analize the following situations:

\begin{itemize}
\item Consider argument $C$ (Figure \ref{argumentos1}) and argument $A$ (Figure \ref{argumentos}). Both arguments have the same claim but different supports. This means that there is a superfluous attack between $C$ and $A$.
\item The above situation is the clearest way for identifying a superfluous attack. The question is: what happens with the sub-arguments of $C$ and $A$? If we have the set of arguments $\{C,A,D,E\}$ and only there is a conflict between $C$ and $A$ it means that the agent can perform, for example, the plans represented by arguments $C,D,$ and $E$, which means that the agent will perform two plans that lead to the same end because both $D$ and $E$ allow the agent to achieve $clean(5,5)$. Therefore, there also should be a superfluous attack between $D$ and $E$. We can conclude that the sub-arguments of two arguments that attack each other by means of a superfluous attack, also attack each other. Nevertheless, we noticed that there is an exception. Suppose that both $C$ and $A$ have a sub-argument $J$, such that $J$ is part of the three of $C$ and is part of the three of $A$. If all the sub-arguments of $C$ attack all the sub-arguments of $A$, this means that $D$ attacks $J$ and $E$ attacks $J$, which leads to an attack between two sub-arguments of the same three. Thus, we finally conclude that all the sub-arguments of $C$ should attack all the sub-arguments of $A$, except those ones that are the same in both threes.
\item We have analized the relation between the sub-arguments of $C$ and $A$; however, we have not analized the relation between $C$ and the sub-arguments of $A$ and vice-verse. The case is similar to the previous analyse. Suppose that there is a superfluous attack between $C$ and $A$ and there are superfluous attacks between their sub-arguments. This means that the agent could perform the plans represented by arguments $C$ and $E$, which is also a superfluous situation. Thus, we conclude that there should be also a superfluous attack between argument $C$ and the sub-arguments of $A$ and vice-verse. In this case, we also consider the exception described in the previous item.

\end{itemize}

Next, we present the definition of superfluous attack taking into consideration the previous analysis.

\begin{defn}\label{defsuperf} \textbf{(Superfluous attack - $\Ra_s$)} Let $A,B \in \Aa rg$ be two arguments. We say that $(A,B) \in \Ra_s$ occurs when either of the following cases hold:

\begin{enumerate}
\item \textit{Case 1:}
\begin{itemize}
\item $\mathtt{CLAIM}(A) = \mathtt{CLAIM}(B)$,
\item $\mathtt{SUPPORT}(A) \neq \mathtt{SUPPORT}(B)$.
\end{itemize}
\item \textit{Case 2:}

\begin{itemize}
\item $\mathtt{CLAIM}(A) \neq \mathtt{CLAIM}(B)$,
\item $\exists A',B' \in \Aa rg$ such that $(A',B') \in \Ra_s$;
\item $[H'',\mathtt{CLAIM}(A)] \in \mathtt{SUPPORT}(A') $ and $[H''',\mathtt{CLAIM}(B)] \in \mathtt{SUPPORT}(B')$.
\end{itemize}

\item \textit{Case 3:}

\begin{itemize}
\item $\mathtt{CLAIM}(A) \neq \mathtt{CLAIM}(B)$,
\item $\exists A' \in \Aa rg$ such that $(A',B) \in \Ra_s$;
\item $[H'',\mathtt{CLAIM}(A)] \in \mathtt{SUPPORT}(A')$,
\item $[H'',\mathtt{CLAIM}(A)] \notin \mathtt{SUPPORT}(B)$.
\end{itemize}

\end{enumerate}
\end{defn}

When there is a superfluous attack between two arguments, we say that the goals in their conclusions are superfluous conflicting goals. Lastly, relation $\Ra_s$ is symmetric.

\begin{prop}\label{propsims} If $(A, B) \in \Ra_s$, then $(B, A) \in \Ra_s$.

\end{prop}

\begin{figure}[!htb]
	\centering
	\includegraphics[width=0.7\textwidth]{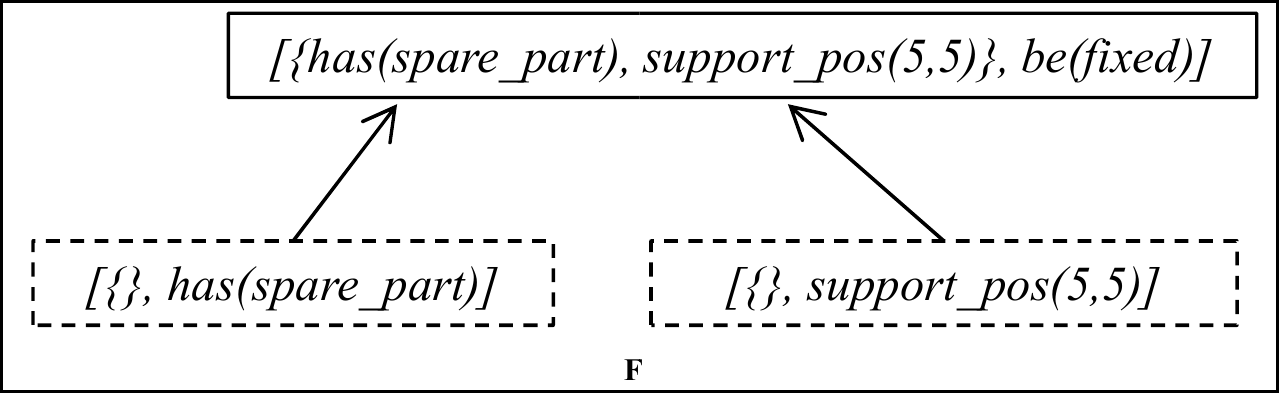} 
	\caption{Argument $F$ for Example \ref{ejemsup}. Dotted-border squares represent the leaves of the three.}
	\label{argumentof}
\end{figure}

\begin{exem}\label{ejemsup}
Consider arguments $A,B,H$ and $E$ (Figure \ref{argumentos}), arguments $C$ and $D$ (Figure \ref{argumentos1}) and argument $F$ (Figure \ref{argumentof}). Arguments $C$ and $A$ and arguments $B$ and $F$ have the same claim namely $clean(5,5)$ and $be(fixed)$, respectively. Hence there is a superfluous attack between $C$ and $A$ and between $B$ and $F$. According to the definition of superfluous attack, this is extended to the sub-arguments of these arguments. Thus, we have that\break $\Ra_s=\{(C,A), (A,C), (E,D), (D,E), (C,E), (E,C), (A,D), (D,A), (F,B), (B,F), (F,H), (H,F)\}$. The attacks that emerge due to Case 1 are: $(C,A), (A,C),(F,B),(B,F)$, the attacks that emerge due to Case 2 are: $(E,D), (D,E)$, and the attacks that emerge due to Case 3 are: $(C,E), (E,C), (A,D), (D,A),(F,H), (H,F) $.

\end{exem}

\section{Postulates concerning Attack Relations}
\label{postulados}
In this section, we study a set of properties that are relevant for the attack relations defined in Section \ref{ataques}. These are based on the postulates presented by Gorogiannis and Hunter \cite{gorogiannis2011instantiating}, which describe desirable properties of the different kinds of attacks that may occur between logical arguments\footnote{You can find the definitions of the main kinds of attacks in \cite{besnard2014constructing}.}. In this work, we have proposed three kinds of attacks -- related to the instrumental arguments -- and these properties are important to guarantee that our approach can be seen as an instance of the abstract argumentation and; therefore, it will benefit from the techniques of abstract argumentation frameworks, explicitly for be able to employ abstract argumentation semantics.

We begin defining the concept of equivalence because it is essential for the understanding most of the properties. We consider three types of equivalence: (i) the logical one, that takes into account the logical structure of the arguments, (ii) the resource equivalence, that considers only the resources, and (iii) the whole equivalence, that takes into account both the logical structure and the resources of an argument.

\begin{defn} \textbf{(Logical equivalence  $\equiv_l$)} Two instrumental arguments $A$ and $B$ are logically equivalent when (i) $\mathtt{CLAIM}(A) = \mathtt{CLAIM}(B)$ and (ii) $\mathtt{SUPPORT}(A)\vdash \mathtt{CLAIM}(B)$ and $\mathtt{SUPPORT}(B) \vdash \mathtt{CLAIM}(A)$ . We denote the logical equivalence by $A \equiv_l B$. 
\end{defn}

In this approach, we can compare two arguments from the point of view of their logical structure or from the point of view of resources necessary for performing the plans represented by the arguments. Next definition states when two arguments are equivalent taking into account their resources.

\begin{defn} \textbf{(Resource equivalence  $\equiv_r$)}\label{resequiv} Two instrumental arguments $A$ and $B$ are  equivalent by resources if $\mathtt{REC}(A)= \mathtt{REC}(B)$. 
We denote the resource equivalence by $A \equiv_r B$. 
\end{defn}

The whole equivalence is defined over the logical equivalence and the resource equivalence definitions.  

\begin{defn} \textbf{(Whole equivalence $\equiv_w$)} Two instrumental arguments $A$ and $B$ are wholly equivalent if (i) they are logically equivalent and (ii) they are equivalent by resources. We denote the whole equivalence by $A \equiv_w B$. 
\end{defn}

From now on, $A$, $B$, $C$ and their primed versions will stand for arguments. Let us recall that $\Ra_t$, $\Ra_r$, and $\Ra_s$ stands for the partial-plans rebuttal, the resource attack, and the superfluous attack, respectively. 

The next six propositions take into account the notion of equivalence to identify attacks that originate due to the equivalent arguments. Proposition \ref{equivter} holds for the partial-plans rebuttal and the superfluous attack, in which only the logical part of the argument is evaluated; therefore, it does not hold for the resource attack. Otherwise, Proposition \ref{equivres} holds for resource attack and Proposition \ref{equivall} holds for any of the types of attacks between arguments. Proposition  \ref{equivabres} mandates that if there is a resource attack between  an argument $A$ and another argument $B$ then there is a resource attack between $B$ and all arguments that are resource equivalent with $A$. In a similar way, Proposition \ref{equivabx} mandates that if there is a partial-plans rebuttal or superfluous attack between arguments $A$ and $B$ then there is a partial-plans rebuttal or superfluous attack, respectively, between $B$ and all arguments that are logically equivalent with $A$. Lastly, Proposition \ref{equivabw} states that if there is a partial-plan (resource or superfluous) attack between arguments $A$ and $B$ then there is a partial-plan (resource or superfluous) attack between $B$ and all arguments that are wholly equivalent with $A$.

\begin{prop} \label{equivter}
If $A \equiv_l A'$ and $B \equiv_l B'$ then $(A,B) \in \Ra_{x} =(A',B') \in \Ra_{x}$ (for $x \in \{t,s\}$).
\end{prop}

\begin{prop}\label{equivres} If $A \equiv_r A'$ and $B \equiv_r B'$ then $(A,B) \in \Ra_r=(A',B') \in \Ra_r$.
\end{prop}

\begin{prop}\label{equivall} If $A \equiv_w A'$ and $B \equiv_w B'$ then $(A,B) \in \Ra_{x}$=$(A',B') \in \Ra_{x}$ (for $x \in \{t,s,r\}$).
\end{prop}

\begin{prop}\label{equivabres} If $(A,B) \in \Ra_r $ and $A \equiv_r  A'$ then $(A',B) \in \Ra_r$.

\end{prop}

\begin{prop}\label{equivabx} If $(A,B) \in \Ra_x$ and $A \equiv_l  A'$ then $(A',B) \in \Ra_x$ (for $x \in \{t,s\}$).

\end{prop}

\begin{prop}\label{equivabw} If $(A,B) \in \Ra_x$ and $A \equiv_w  A'$ then $(A',B) \in \Ra_x$, for $x \in \{t,r,s\}$.

\end{prop}

The following proposition holds for the partial-plans rebuttal. It means that when two partial plans of different instrumental arguments have an inconsistency in their heads, then both arguments attack each other. Notice that this postulate is only related to partial-plan attack.

\begin{prop}\label{propsinnum1} If $(A,B) \in \Ra_t$, then  $\exists [H,\psi] \in \mathtt{SUPPORT}(A)$ and $\exists [H',\psi'] \in \mathtt{SUPPORT}(B)$ such that $\mathtt{HEAD}([H,\psi]) \cup \mathtt{HEAD}([H',\psi']) \vdash \perp$.
\end{prop}

\section{Argumentation Frameworks}
\label{frames}

In this section, we present an argumentation framework for each kind of incompatibility (i.e., terminal, resource, and superfluity) and a general argumentation framework that involves all the of arguments and attacks of the three kinds of incompatibility.

\begin{defn} \textbf{(Argumentation framework)} Let $\Aa rg $ be the set of arguments that can be built from the agent's $\Ka\Ba$. 

\begin{itemize}
\item A $x$-AF is a pair $\Aa\Fa_x=\langle \Aa rg_x, \Ra_x  \rangle$ (for $x \in \{t,r,s\}$) where $\Aa rg_x \subseteq \Aa rg$ and $\Ra_x$ is the binary relation in $\Aa rg_x$.
\item A general g-AF is a pair $\Aa\Fa_g=\langle \Aa rg, \Ra_g \rangle$, where $\Ra_g= \Ra_t \cup \Ra_r \cup \Ra_s$.

\end{itemize}

\end{defn}

Notice that it may occur that there exist the three kinds of attacks between two arguments. In this case, we consider multiple attacks between two arguments as a unique general attack.

\begin{exem} \label{ejmAFtotal} Considering the arguments and attacks of the three kinds of incompatibility, we have the following g-AF:  $\Aa\Fa_g=\langle \{A,B,C,D, E, F,H\},\{ (A,B), (B,A), (E,B), (B,E), (E,H),(H,E),(A,H),(H,A), (C,B),(B,C),\break (D,B),(B,D), (D,H), (H,D), (C,H),(H,C),(C,A), (A,C),(E,D), (D,E), (C,E), (E,C), (A,D), (D,A),\break (F,B), (B,F), (F,H), (H,F) \} \rangle$. Figure \ref{aftotal} show the graph representation of this framework, where nodes represent the arguments and edges the attacks.

\begin{figure}[!h]
	\centering
	\includegraphics[width=0.8\textwidth]{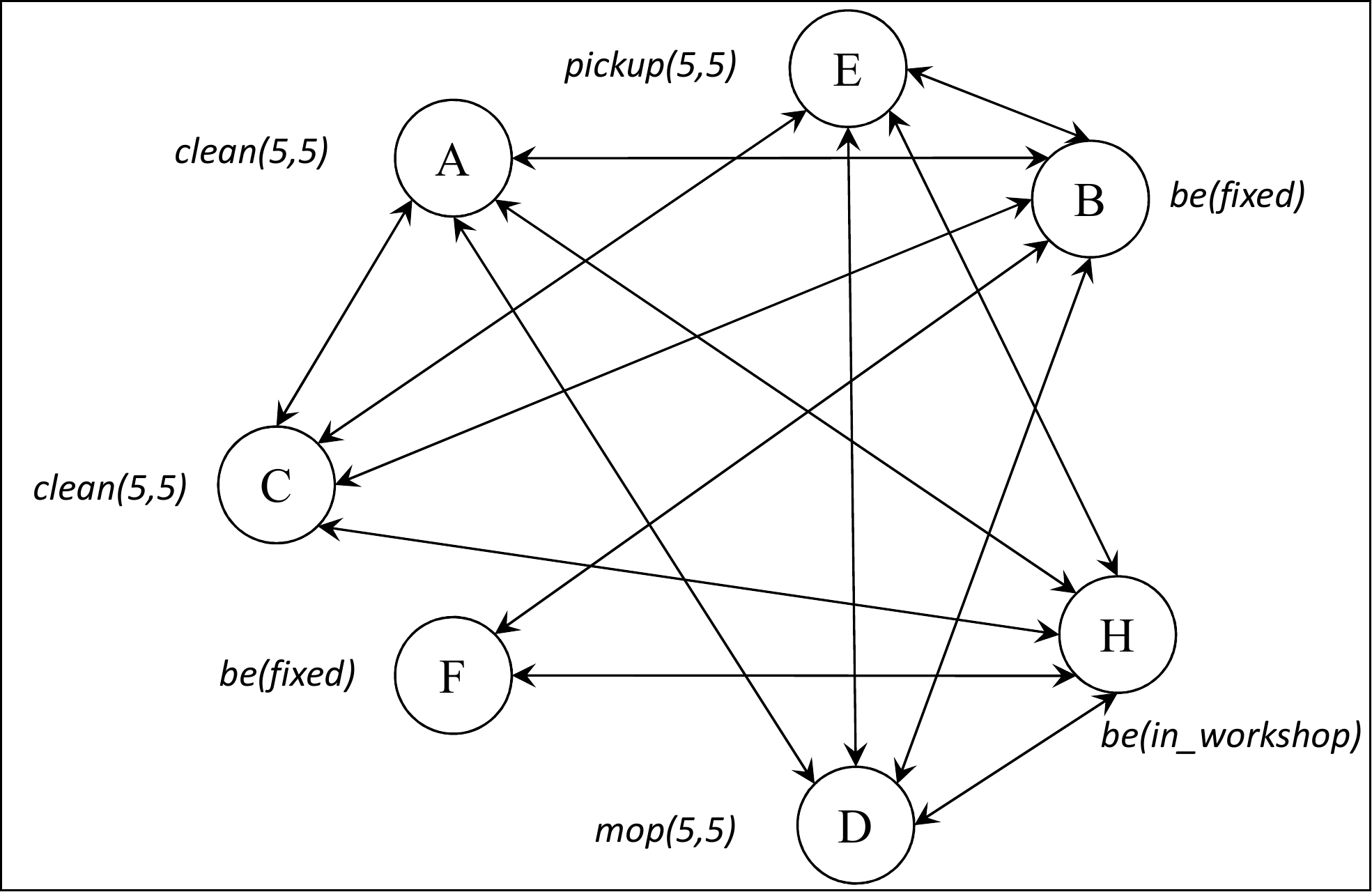} 
	\caption{The general argumentation framework $\Aa\Fa_g$ for Example \ref{ejmAFtotal}. The nodes represent the arguments that are part of $\Aa\Fa_g$ and the arrows represent the attacks between the arguments. The text next to each node indicates the claim of each argument.}
	\label{aftotal}
\end{figure}
\end{exem}

Hitherto, we have considered that all attacks are symmetrical. However, as mentioned in the introduction section and in Section \ref{bloques}, goals have a preference value, which indicates how valuable each goal is for the agent. Since plans allow the agent to achieve his goals, we can say that they can inherit the preference value of the corresponding goal. In other words, instrumental arguments inherit the preference value of the goal in their claims. Therefore, depending on this preference value, which is returned by function $\mathtt{PREF}$, some attacks may be considered successful. This means that the symmetry of the relation attack may be broken.

\begin{defn} \textbf{(Successful attack)}\footnote{In other works, it is called a defeat relation \cite{martinez2006progressive} \cite{modgil2014aspic+}.} Let $A,B \in \Aa rg$ be two arguments, we say that $A$ successfully attacks $B$ when $(A,B) \in \Ra_g$ and $\mathtt{PREF}(\mathtt{CLAIM}(A)) > \mathtt{PREF}(\mathtt{CLAIM}(B))$.

Let us denote with $\Aa\Fa_g'$ the general argumentation framework that results after considering the successful attacks.
\end{defn}

\begin{exem}\label{ejmdefeat} (Cont. Example \ref{ejmAFtotal}) Consider the following preference values for the goals that are the claims of the arguments: $\mathtt{PREF}(clean(5,5))=0.75$,  $\mathtt{PREF}(be(fixed))=0.6$, $\mathtt{PREF}(mop(5,5))=0.8$, $\mathtt{PREF}(pickup(5,5))=0.75$, and $\mathtt{PREF}(be(in\_workshop))=0.6$. The general AF after considering the successful attacks is $\Aa\Fa_g'=\langle \{A,B,C,D, E, F,H\},\{ (A,B), (E,B), (E,H),(A,H), (C,B), (D,B), (D,H),  (C,H),(C,A), (A,C),(D,E),\break (C,E), (E,C), (D,A), (F,B), (B,F),(F,H), (H,F) \} \rangle$. Figure \ref{figdefeat} shows the graphical representation of $\Aa\Fa_g'$.

\begin{figure}[!htb]
	\centering
	\includegraphics[width=0.8\textwidth]{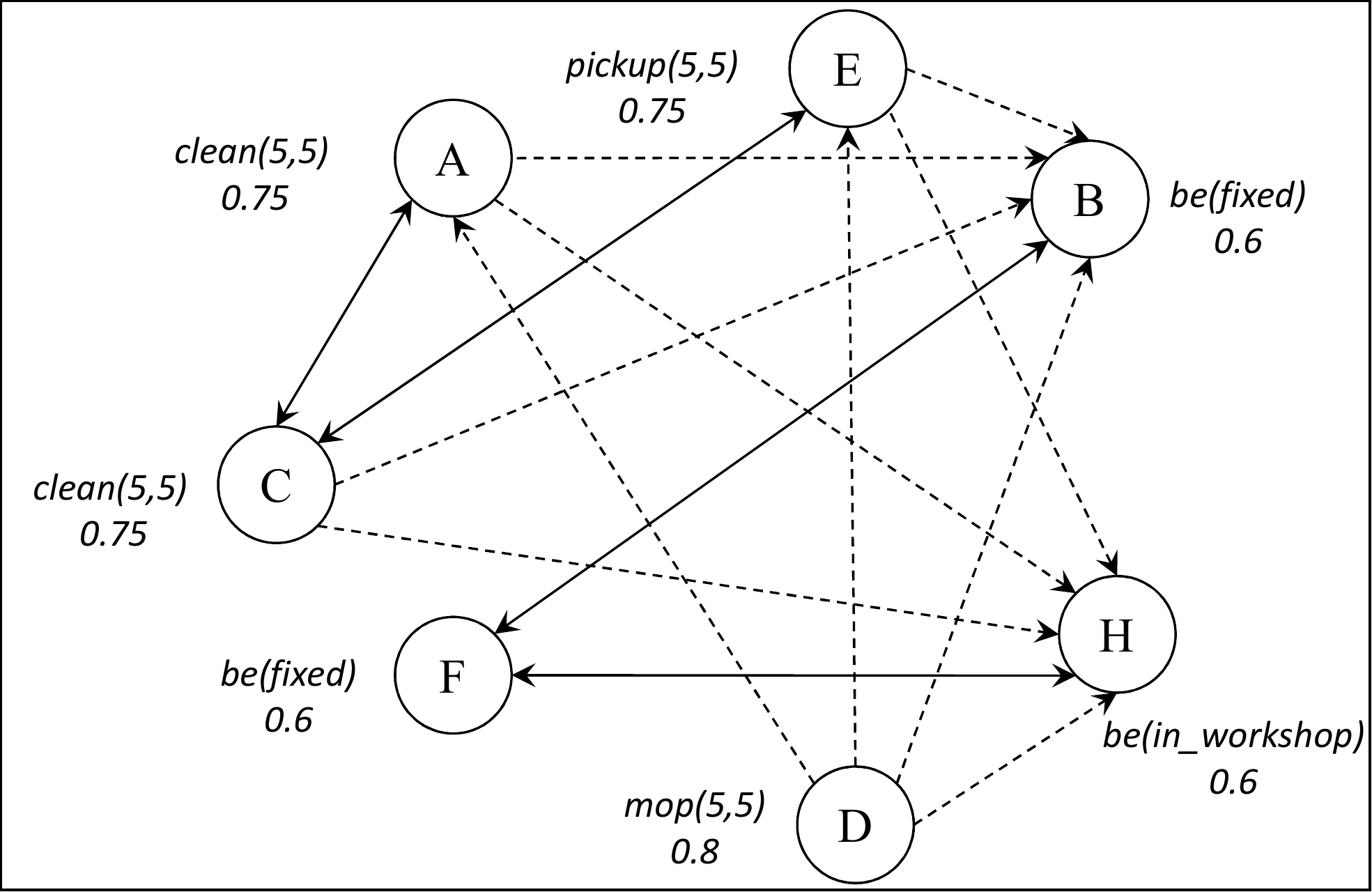} 
	\caption{The general argumentation framework $\Aa\Fa_g'$ after considering the preference value. Dashed-line arrows represent the successful attacks. Next to each node there is information about the claim of the argument and the preference value of it.}
	\label{figdefeat}
\end{figure}

\end{exem}

\section{Selection of Compatible Goals}
\label{seleccion}
In this section, we use the general argumentation framework $\Aa\Fa_g'$, which takes into account the successful attacks, in order to select those goals that can be considered compatible. While it is true that we defined an AF for each form incompatibility, we will use the general AF because it includes all the kinds of attacks that exist between the arguments and this is important for a proper identification of incompatible goals. However, depending on the application or the interests of the agent, he could use one or two incompatibility AF instead of the general one. The process of selection will be the same for any case. 

Argumentation is related to non-monotonic reasoning. This reasoning allows to retract the inferences based on new information. This kind of reasoning is important in the context of goal reasoning because new information about the plans can change the attack relation and therefore the set of compatible goals. Additionaly, for the selection of the compatible goals we use the so called argumentation semantics \cite{dung1995acceptability}, which are ideal to determine non-conflicting elements out of a set of conflicting ones without considering the nature of the conflict. In this work, we have formalized three types of incompatibilities of different nature. Related work has also deal with these incompatibilities, but in a separate way, that is, by means of different approaches (see Related Work in Section \ref{correlatos}). Therefore, there is a need of a unique approach that integrates and resolves the different types of incompatibilities, and we can do that by using argumentation semantics.

We propose to apply argumentation semantics in two different ways: (i) we apply a semantics on the instrumental arguments and the attacks that belong to the general AF, and (ii) we apply a semantics on a new argumentation framework that is made up of pursuable goals and the attacks between them. This argumentation framework results by identifying when two goals are incompatible from the general AF.

\subsection{Semantics for the selection process}
\label{semantica}

So far we have defined instrumental arguments, which represent plans, and we have introduced the kinds of attacks that determine each form of incompatibility. The next step is to determine the set of goals that can be executed without conflicts, which can also be called acceptable goals.  This will be done by applying argumentation semantics. Let us present an empirical analysis that allows us to determine what argumentation semantics concepts should be taken into account. 

\begin{figure}[!htb]
	\centering
	\includegraphics[width=0.9\textwidth]{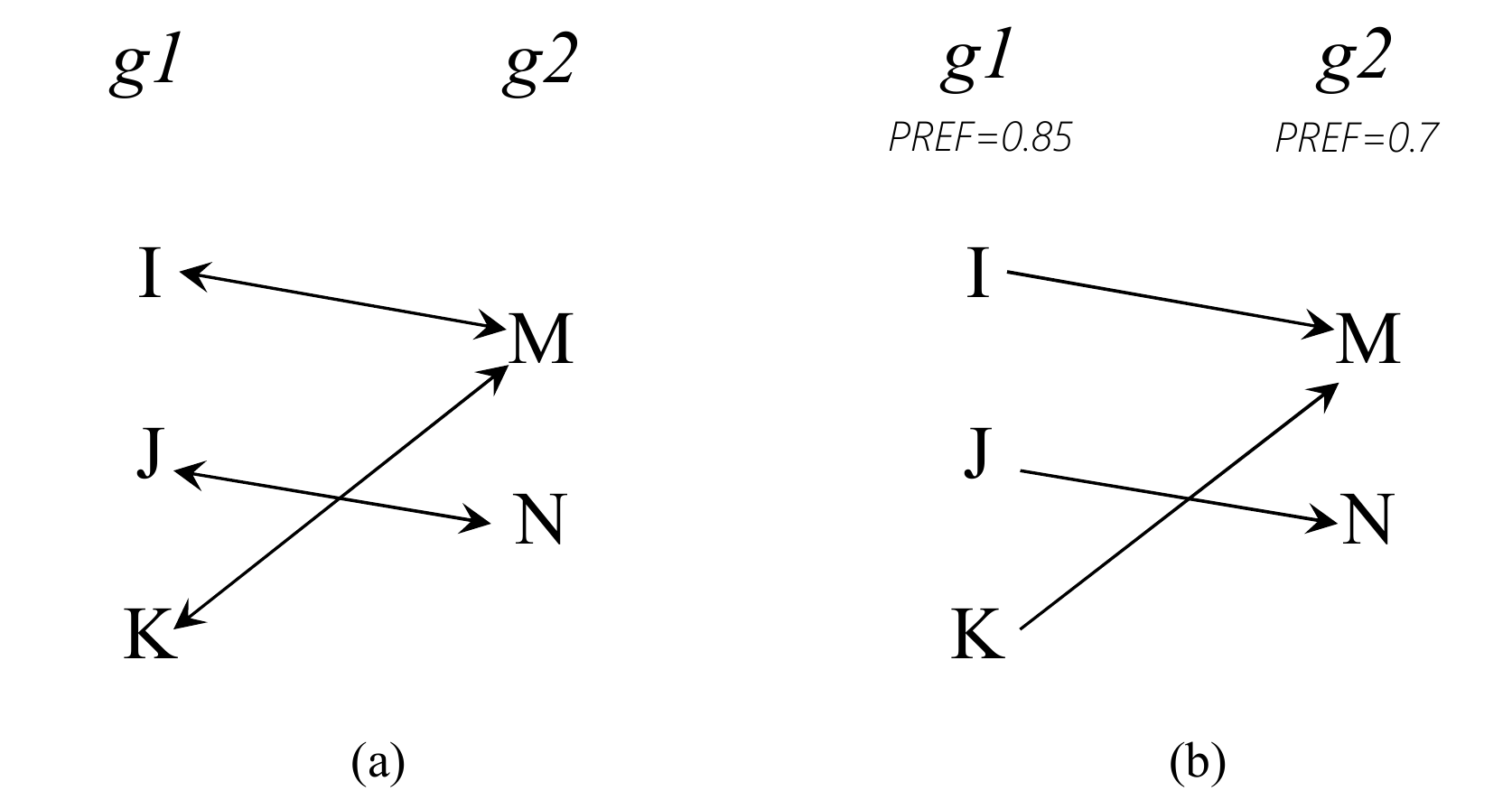} 
	\caption{(a) The general argumentation framework before considering the successful attacks. (b) The general argumentation framework after considering the successful attacks.
Notice that the attack relation is no longer symmetrical.}
	\label{analisissim}
\end{figure}

For this analysis, we will use a simpler AF than the one that has been generated throughout the article. Figure \ref{analisissim}(a) shows the general AF generated for goals $g1$ and $g2$ and Figure \ref{analisissim}(b) shows the general AF after considering the successful attacks. Goal $g1$ is the claim of arguments $I,J,$ and $K$, and goal $g2$ is the claim of arguments $M$ and $N$. Note that there is no sub-arguments in this structure.

\begin{enumerate}
\item Consider Figure \ref{analisissim}(b). Under the preferred semantics (i.e. an admissibility-based semantics) and the stage semantics (i.e. a semantics based on conflict-freeness and the range notion), the returned set is $\{I,J,K\}$. This would mean that the selected goal is $g1$; however, this is not proper\footnote{$\{J, M\}$ or $\{I, N\}$ are conflict-free sets that can be considered proper extensions since they allow the agent to achieve both goals $g1$ and $g2$. Thus, a conflict-free extension $\Ea $is not proper with respect to other conflict-free extensions when there is at least other conflict-free extension that allows the agent to achieve more goals than $\Ea$.}. 

\item Consider now Figure \ref{analisissim}(a). Under the preferred and the stage semantics, we obtain $\{I,J,K\},\{I,K,N\}, \{J,M\}, \{M,N\}$. The question is: which of these extensions determines the selected goals? If we take $\{I,J,K\}$ or $\{M,N\}$, it would mean that only one of the goals will be selected. On the other hand, if we take $\{I,K,N\}$ or $\{J,M\}$, it would mean that both goals will be selected. Therefore, there should be a function that supports the agent to decide which extension to choose faced with more than one extension. 

\end{enumerate}

In summary, we can observe that (i) when the notion of successful attacks is not considered, the semantics may not return the proper extension (for the simpler example it was not returned) and (ii) when the notion of successful attacks is considered, the semantics return at least one proper extension. In this last case, the agent has to choose an extension. 

Regarding to the first analyzed situation, note that none of the semantics returns a proper extension for the simpler example. However, if we only apply the conflict-free semantics, there are some proper extensions that contain arguments associated to both goals. Based on this, we can also say that:
\begin{itemize}
\item The concept of defence is not necessary. Observe extension $\{I, N\}$, which is a proper extension. In this extension, argument $N$ is attacked by argument $J$ and argument $I$ does not defend it. Even so, this extension continuous being proper. It does not mean that there could be cases in which arguments of an extension defend other arguments; however, it is not determinant.
\item The number of elements of an extension does not determine the number of compatible goals. For example, extension $\{I,J,K\}$ has more elements than extension $\{I,N\}$, but the last one leads to obtain a set of two compatible goals, and the first one contains arguments for only one goal. Hence, the conflict-free extensions with more amount of elements does not always are the proper ones.
\end{itemize}

Now, the question is: which extensions can be considered the proper ones? There are two possible criteria for determining it: (i) the maximum number of compatible pursuable goals and (ii) the maximum total preference value taken into account only pursuable goals. Thus, the agent may choose whether he wants to commit to the maximum number of pursuable goals or he wants to commit to those pursuable goals that maximize the preference value.

Thus, for determining the proper extension(s), we consider the following steps:

\begin{enumerate}

\item Apply the conflict-free semantics to the g-AF resultant after considering the successful attacks, i.e. to $\Aa\Fa_g'$.

\item If the agent wants first to maximize the number of compatible pursuable goals, the extension(s) that fulfils this requisite should be selected. If there is more than one extension, then the agent should select the extension(s) that maximizes the sum of the preference value of the goals.

\item If the agent wants first to maximize the total preference value, the extension(s) that fulfils this requisite should be selected. If there is more than one extension, then the agent should select the extension(s) that maximizes the number of compatible pursuable goals.
\item Finally, the agent selects those extensions that are maximal with respect to the set inclusion.
\end{enumerate}

\begin{defn} \textbf{(Semantics functions)} Given a general argumentation framework $\Aa\Fa_g'=\langle \Aa rg, \Ra_g \rangle$. Let $\Sa_{\Ca\Fa}$ be a set of conflict-free sets calculated from $\Aa\Fa$. The following functions are defined to determine the set of proper extensions:

\begin{itemize}
\item $\mathtt{MAX\_GOAL}: \Sa_{\Ca\Fa} \rightarrow 2^{\Sa_{\Ca\Fa}}$. This function takes as input a set of conflict-free sets and returns those sets with the greatest number of pursuable goals. Sub-goals are not taken in to account in this function.

\item $\mathtt{MAX\_UTIL}: \Sa_{\Ca\Fa} \rightarrow  2^{\Sa_{\Ca\Fa}}$. This function takes as input a set of conflict-free sets and returns those with the maximum utility for the agent in terms of preference value. The utility of each extension is calculated by summing up the preference value of all the goals, i.e. pursuable goals and sub-goals.  

\item $\mathtt{COMP\_GOALS}: \Sa_{\Ca\Fa}^p \rightarrow  2^{\Ga_p}$. This function takes as input a proper extension and returns a set of compatible goals.  Elements of $\Sa_{\Ca\Fa}^p$ are proper extensions that are returned by the application of the above functions. 
\end{itemize}
\end{defn}

Algorithm \ref{evalincomp} shows the steps for determining the sets of compatible goals. The output of the algorithm is a set of sets because more than one proper extension can be returned by the above functions. The algorithm includes the steps for determining the set of proper extensions.

\begin{algorithm}
\label{evalincomp}
\begin{algorithmic}[1]
\REQUIRE The general argumentation framework $\Aa\Fa_g'$
\ENSURE Sets of compatible goals $\Sa_{comp}$
\STATE{$\Sa_{comp}$=\{\}}
\STATE{\textit{/*--------------- Steps to determine the proper extensions ---------------*/}}
\STATE{$\Sa_{\Ca\Fa}=\mathtt{CONFLICT-FREE}(\Aa\Fa_g')$ \textit{//Function }$\mathtt{CONFLICT-FREE}$ \textit{returns the conflict-free sets of an AF}} 
\IF{First maximize the number of compatible goals}
	\STATE{$\Sa_{\Ca\Fa}'=\mathtt{MAX\_GOAL}(\Sa_{\Ca\Fa})$}
	\IF{$|\Sa_{\Ca\Fa}'|>1$}
		\STATE{$\Sa_{\Ca\Fa}''=\mathtt{MAX\_UTIL}(\Sa_{\Ca\Fa'})$}
	\ENDIF
\ELSIF{First maximize the total preference value}
	\STATE{$\Sa_{\Ca\Fa}'=\mathtt{MAX\_UTIL}(\Sa_{\Ca\Fa})$}
	\IF{$|\Sa_{\Ca\Fa}'|>1$}
		\STATE{$\Sa_{\Ca\Fa}''=\mathtt{MAX\_GOAL}(\Sa_{\Ca\Fa'})$}
	\ENDIF
\ENDIF
\STATE{$\Sa_{\Ca\Fa}^p= \mathtt{MAXIMAL}(\Sa_{\Ca\Fa}'')$ \textit{//Function} $\mathtt{MAXIMAL}$ \textit{returns the extensions that are maximal with respect to the set inclusion}} 
\STATE{\textit{/*--------- End of the steps to determine the proper extensions ---------*/}}
\STATE{\textit{/*--------- Final step to determine the sets of compatibles goals ---------*/}}
\STATE{$n=|\Sa_{\Ca\Fa}^p|$}
\FOR {$count=1$ to n}
	\STATE{$\Ga_p'=\mathtt{COMP\_GOALS}(\Ea_{count}')$}
	\STATE{$\Sa_{comp}=\Sa_{comp} \cup \Ga_p'$}
\ENDFOR
\RETURN {$\Sa_{comp}$}
\end{algorithmic}
\caption{Steps for determining the sets of compatible goals}
\label{evalincomp}
\end{algorithm}

\begin{exem}\label{ejmplanos} (Cont. Example \ref{ejmdefeat}) Recall that $\Aa\Fa_g'=\langle \{A,B,C,D, E, F,H\}, \{ (A,B), (E,B), (E,H),(A,H), (C,B),\break (D,B), (D,H),  (C,H), (C,A), (A,C),(D,E), (C,E), (E,C),  (D,A), (F,B), (B,F), (F,H), (H,F) \} \rangle$ (Figure \ref{figdefeat}). Recall also that $\mathtt{PREF}(clean(5,5))=0.75$,  $\mathtt{PREF}(be(fixed))=0.6$,\break $\mathtt{PREF}(mop(5,5))=0.8$, $\mathtt{PREF}(pickup(5,5))=0.75$, and $\mathtt{PREF}(be(in\_workshop))=0.6$. Table 1 shows all the conflict-free sets $\Sa_{\Ca\Fa}$, the number of compatible pursuable goals of each set and the utility of each set. 

\begin{table}[!htb]
\label{tbejmplanos}
\begin{center}
\begin{tabular}{|c|c|c|c|c|c|c|}
\hline 
{\scriptsize EXTENSION} & {\scriptsize NUM. GOALS} & {\scriptsize UTILITY} &  & {\scriptsize EXTENSION} & {\scriptsize NUM. GOALS} & {\scriptsize UTILITY} \\ 
\hline 
$\{\}$ & 0 & 0 & & $\{D,F\}$ & 1 & 1.4 \\ 
\hline 
$\{F\}$ & 1 & 0.6 &  & $\{C\}$ & 1 & 0.75 \\ 
\hline 
$\{E\}$ & 0 & 0.75 &  & $\{C,F\}$ & 2 & 1.35 \\ 
\hline 
$\{E,F\}$ & 1 & 1.35&  & $\{C,D\}$ & 1 & 1.55 \\ 
\hline 
$\{A\}$ & 1 & 0.75 &  & $\{C,D,F\}$ & 2 & 2.15 \\ 
\hline 
$\{A,F\}$ & 2 & 1.35 &  & $\{H\}$ & 0 & 0.6 \\ 
\hline 
$\{A,E\}$ & 1 & 0.75 &  & $\{B\}$ & 1 & 0.6 \\ 
\hline 
$\{A,E,F\}$ & 2 & 2.1 &  & $\{B,H\}$ & 1 & 1.2 \\ 
\hline 
$\{D\}$ & 0 & 0.8 &  &  &  &  \\ 

\hline 
\end{tabular}

\end{center}\caption{Conflict-free sets for Example \ref{ejmplanos}. Column {\scriptsize NUM. GOALS} corresponds to the number of compatible pursuable goals the plans (represented by arguments) of each extension allow to achieve. Column {\scriptsize UTILITY} shows the total preference value summing up all the goals of the extension.}
\end{table}

In this case, the decision of the agent about to first maximize the number of goals or first maximize the utility does not affect the final result. Thus, $\mathtt{MAX\_GOAL} \circ \mathtt{MAX\_UTIL}(\Sa_{\Ca\Fa})=  \mathtt{MAX\_UTIL} \circ \mathtt{MAX\_GOAL}= \{\{C, D,F\}\}$. This extension projects three compatible goals, of which two are pursuable goals, with an utility of 2.15. Let's see now the set of projected compatible goals: $\mathtt{COMP\_GOALS}(\{C, D,F\})= \{clean(5,5), be(fixed), mop(5,5)\}$, where $clean(5,5)$ and $be(fixed)$ are pursuable goals and $mop(5,5)$ is the sub-goal of $clean(5,5)$.
\end{exem}

Next example shows that the order of application of the semantics functions may affect the final result.

\begin{exem} Given a general argumentation framework $\Aa\Fa_g''=\langle \{A,B,C\}, \{(A,B), (A,C)\} \rangle$. Goal $g1$ is the claim of argument $A$, goal $g2$ is the claim of argument $B$, goal $g3$ is the claim of argument $C$. Consider $\mathtt{PREF}(g1)=0.9$, $\mathtt{PREF}(g2)=0.3$, and $\mathtt{PREF}(g3)=0.4$. The set of conflict-free extensions is: $\Sa_{\Ca\Fa}=\{\{{\}}, \{C\}, \{B\},\{B,C\},\{A\}\}$.

Case the agent wants to first maximize the number of compatible goals, the result is: $\mathtt{MAX\_GOAL}(\Sa_{\Ca\Fa})=\{B,C\}$. Since there is only one extension, there is no necessity of applying the function $\mathtt{MAX\_UTIL}$. Finally, $\mathtt{COMP\_GOALS}(\{B,C\})=\{g2,g3\}$.

Case the agent wants to first maximize the utility, the result is: $\mathtt{MAX\_UTIL}(\Sa_{\Ca\Fa})=\{A\}$. Since there is only one extension, there is no necessity of applying the function $\mathtt{MAX\_GOAL}$. In this case, the final set of compatible goals is: $\mathtt{COMP\_GOALS}(\{A\})=\{g1\}$.

\end{exem}

\subsection{Selection at goals level}

Another way to obtain the set of compatible goals is by generating a new argumentation framework whose elements are goals. In order to generate this framework, we need to define when two goals attack each other. We make this definition based on the general attack relation $\Ra_g$, which includes the three kinds of attacks that may exist between arguments. Thus, a goal $g$ attacks another goal $g'$ when all the arguments for $g$ have a general attack relation with all the arguments for $g'$. This attack relation between goals is captured by the binary relation $\Ra\Ga \subseteq \Ga' \times \Ga'$ where, given an $\Aa\Fa_g'=\langle \Aa rg, \Ra_g \rangle$, $\Ga'= \bigcup_{A \in \Aa rg} \mathtt{BODY}(\mathtt{SUPPORT}(A)) \cap \Ga$, i.e. $\Ga'$ includes all the pursuable goals and all their sub-goals. We denote with $(g, g')$ the attack relation between goals $g$ and $g'$. In other words, if $(g,g') \in \Ra\Ga$ means that goal $g$ attacks goal $g'$.

\begin{defn} \label{generalinc}\textbf{(Attack between goals)} Let $\Aa\Fa_g'=\langle \Aa rg, \Ra_g  \rangle$ be a general AF, $g, g' \in \Ga'$ be two goals, $\mathtt{ARG}(g),\mathtt{ARG}(g') \subseteq \Aa rg$ be the set of arguments for $g$ and $g'$, respectively. Goal $g$ attacks goal $g'$ when $\forall A \in \mathtt{ARG}(g)$ and $\forall A' \in \mathtt{ARG}(g')$ it holds that $(A,A') \in \Ra_g$ or $(A',A) \in \Ra_g$. 

\end{defn}

This definition states that although there are attack relations between arguments associated to different goals, if there is at least two arguments (one for each goal) that do not have an attack relation, then the goals are considered non-conflicting since there is a way of achieving them. Next proposition formalizes this idea.

\begin{prop}\label{propsinnum2} Let $g,g' \in \Ga'$, $\mathtt{ARG}(g)$ and $\mathtt{ARG}(g')$ be the sets of arguments whose claims are $g$ and $g'$, respectively. If $\exists A \in \mathtt{ARG}(g)$ and $\exists B \in \mathtt{ARG}(g')$ such that $A$ and $B$ are conflict-free, then $(g,g'),(g',g) \not\in \Ra\Ga$.

\end{prop}

\begin{figure}[!htb]
	\centering
	\includegraphics[width=0.6\textwidth]{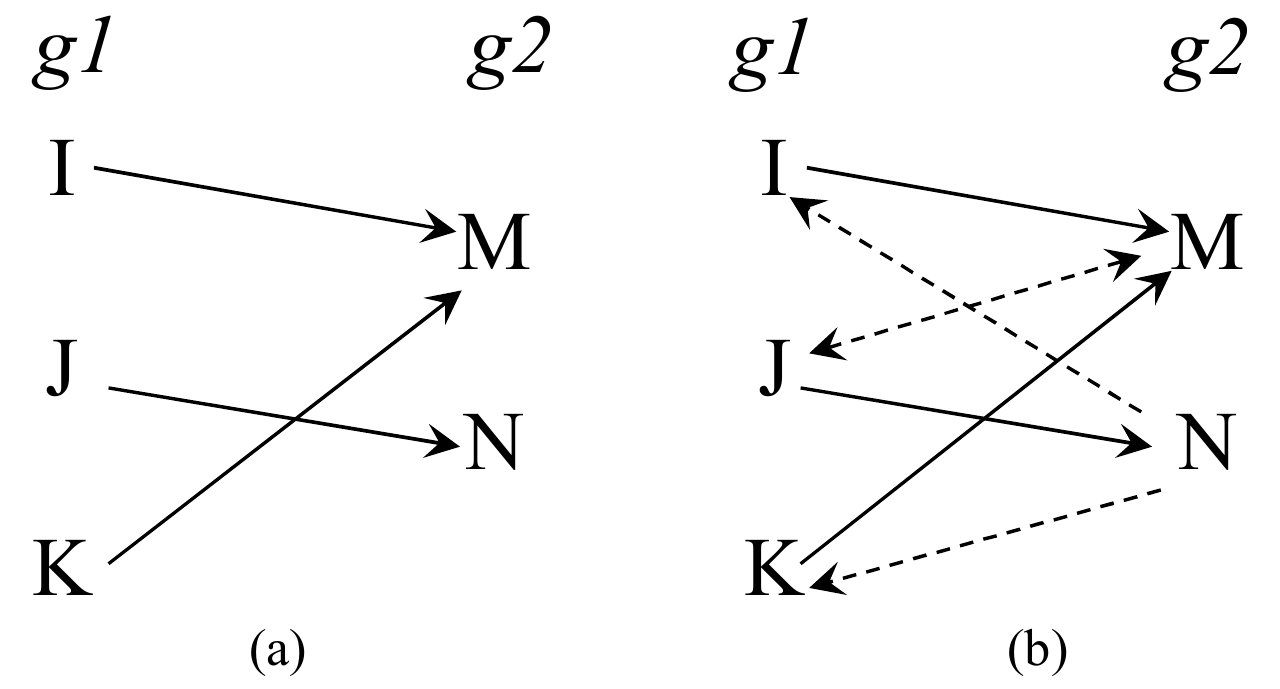} 
	\caption{(a) There is not an attack between goals $g1$ and $g2$, black arrows represent partial-plan rebuttals. (b) Dashed-line arrows represent resource attacks. Notice that partial-plans rebuttal complements resource attacks. Therefore, there is an attack between goals $g1$ and $g2$.}
	\label{complement}
\end{figure}

It is important to notice that for defining the attack between goals the three kinds of attacks between arguments are taken into account. This means that the attacks between arguments can complement each other to generate an attack between goals. Let us explain it graphically. Consider Figure \ref{complement}(a), suppose that black arrows represent partial-plan rebuttals between goals $g1$ and $g2$. Taking into account these attacks between arguments, there is not an attack between goals $g1$ and $g2$. Consider now Figure \ref{complement}(b), besides the partial-plans rebuttal represented by black arrows, suppose that green arrows represent resource attacks. Notice that in Figure \ref{complement}(b), there is an attack between all the arguments for $g1$ and all the arguments for $g2$. In conclusion, we can say that an attack relation between two goals is possible even when the attacks between their arguments are of different types. This means that the attack relation between goals generalizes the attack between arguments.

Now, we define the argumentation framework whose elements are goals. The attack relation between such goals is ruled by Definition \ref{generalinc}.

\begin{defn} \textbf{(Goals argumentation framework)} An argumentation-like framework for dealing with incompatibility between goals is a triple $\Ga\Aa\Fa =\langle \Ga', \Ra\Ga ,\mathtt{PREF} \rangle$, where:

\begin{itemize}
\item $\Ga_p$ is a set of the pursuable goals,
\item $\Ra\Ga\subseteq  \Ga' \times \Ga'$,
\item $\mathtt{PREF}$ is the function that returns the value of a given goal for the agent.
\end{itemize}
 
\end{defn}

The steps for obtaining the proper extensions can also be applied on this framework. However, the application of function $\mathtt{MAX\_GOAL}$ is different since it will return the extension(s) with the most amount of elements. Finally, function $\mathtt{COMP\_GOALS}$ is not necessary in this approach.

\begin{exem}\label{ejemincom} Consider the final general AF of Example \ref{ejmdefeat}. $\Aa\Fa_g'=\langle \{A,B, C,D, E, F,H\},\{ (A,B), (E,B),\break (E,H),(A,H), (C,B), (D,B), (D,H),(C,H), (C,A), (A,C), (D,E), (C,E), (E,C),  (D,A), (F,B), (B,F),\break (F,H), (H,F) \} \rangle$. From $\Aa\Fa$ the agent generates the following AF: $\Ga\Aa\Fa=\langle \{clean(5,5), pickup(5,5), mop(5,5), be(in\_workshop), be(fixed)\}, \{(mop(5,5), pickup(5,5)),(clean(5,5),\break be(in\_workshop)),  (mop(5,5), be(in\_workshop)), (pickup(5,5), be(in\_workshop))\}, \{\mathtt{PREF}(clean(5,5)),\break \mathtt{PREF}(mop(5,5)), \mathtt{PREF}(pickup(5,5)), \mathtt{PREF}(be(in\_workshop)), \mathtt{PREF}(be(fixed))\}  \rangle$ (Figure \ref{framegoals}). Notice that $(clean(5,5), be(in\_workshop)) \in \Ra\Ga$ due to $(A,H),(C,H) \in \Ra_g$, $(mop(5,5), pickup(5,5)) \in \Ra\Ga$ due to $(D,E) \in \Ra_g$, $(pickup(5,5),be(in\_workshop)) \in \Ra\Ga$ due to $(E,H) \in \Ra_g$, and $(mop(5,5),be(in\_workshop)) \in \Ra\Ga$ due to $(D,H) \in \Ra_g$.

\begin{figure}[!h]
	\centering
	\includegraphics[width=0.75\textwidth]{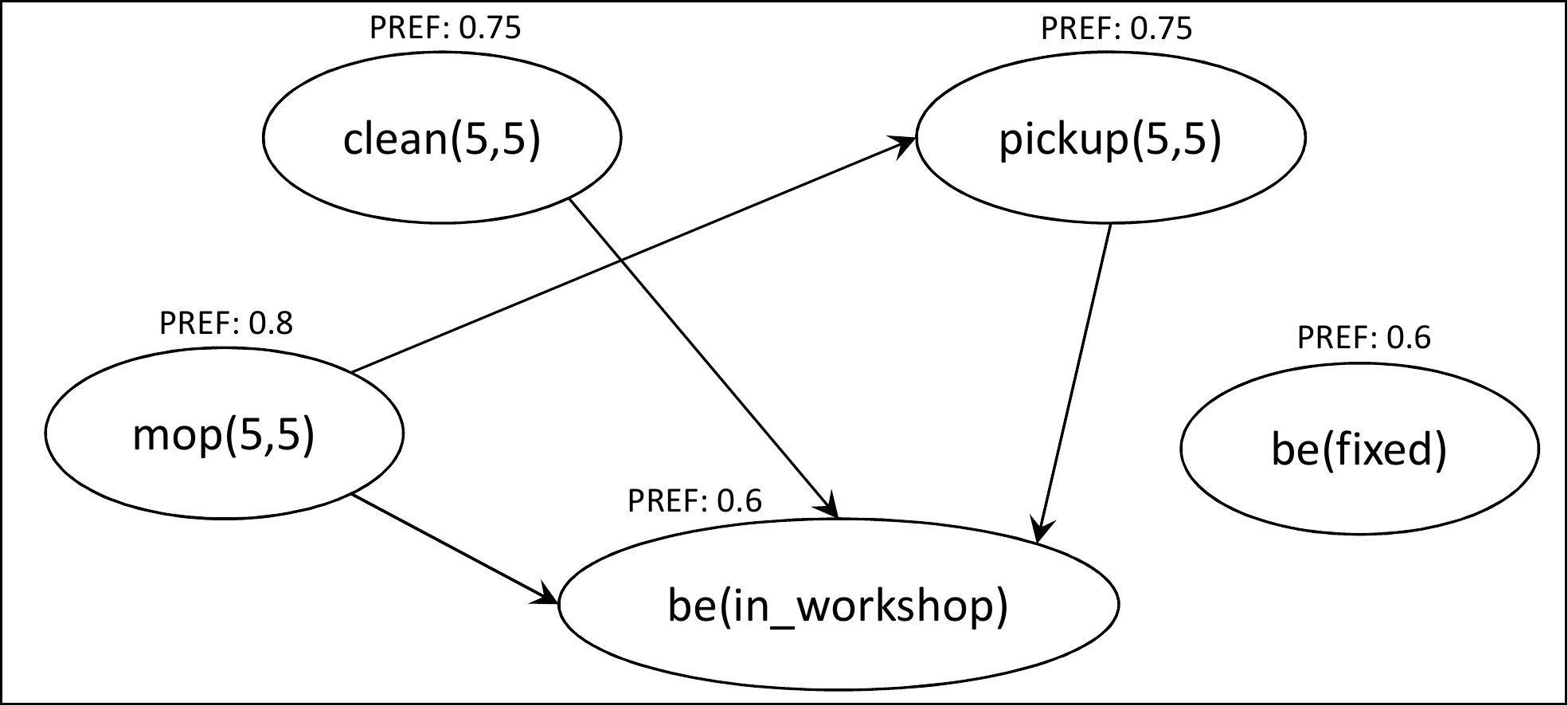} 
	\caption{Graph of the goals argumentation framework of Example \ref{ejemincom}.}
	\label{framegoals}
\end{figure}

The number of conflict-free extensions is: $|\Sa_{\Ca\Fa}|= 14$. By applying $\mathtt{MAX\_GOAL}$, we obtain: $\mathtt{MAX\_GOAL}(\Sa_{\Ca\Fa})= \{\{clean(5,5), mop(5,5), be(fixed)\}, \{clean(5,5), pickup(5,5), be(fixed)\}\}$. The total preference value of the first extension is 2.15 and of the second one is 2.1. Thus, after applying $\mathtt{MAX\_UTIL}$, the selected extension is: $\{clean(5,5), mop(5,5), be(fixed)\}$. This means that goals $clean(5,5)$, $mop(5,5)$, and $be(fixed)$ are compatible and con be executed without conflicts.

\end{exem}

\subsection{Discussion}

Notice that according to the decision of the agent, he could use first $\mathtt{MAX\_GOAL}$ and then $\mathtt{MAX\_UTIL}$ or vice-versa. What must be clear is that the first function to be applied takes as input all the conflict-free sets that are calculated from the general AF and the second function takes as input the sets returned by the first function. Also notice that function $\mathtt{MAX\_GOAL}$ only takes into account pursuable goals, unlike function $\mathtt{MAX\_UTIL}$ that takes into account all the goals of an extension. In order to explain the reason we decided to restrict function $\mathtt{MAX\_GOAL}$ to pursuable goals, let us present the following situation: suppose that $\Ga_p=\{g1,g2\}$ is the set of pursuable goals and $\Ea_1=\{A,B,C,D\}, \Ea_2=\{E,F\}$ are the extensions returned by the conflict-free semantics. We have that $\mathtt{ARG}(g1)=\{A,E\}$, $\mathtt{ARG}(g2)=\{F\}$, and goals $B, C,$ and $D$ are sub-goals of $A$. If we take into account all the goals in an extension, then we can say that $\Ea_1$ allows the agent to achieve more amount of goals; however, only one of them is a pursuable goal. On the other hand, if we take into account only pursuable goals, then we can say that $\Ea_2$ allows the agent to achieve two pursuable goals, i.e. more pursuable goals than $\Ea_1$. We think that it is more important to maximize the amount of pursuable goals than the amount of goals including sub-goals.

\section{Evaluation of the Approach}
\label{evalua}
In this section, we aim to study the properties of the proposed argumentation-based approach. We evaluate this approach and prove that it satisfies the rationality postulates proposed in \cite{caminada2007evaluation}.

Firstly, let us define the following notation:

\hspace{2cm}$\mathtt{BEL}(\Ea)= \bigcup_{A \in \Ea} (\bigcup_{pp \in \mathtt{SUPPORT}(A)}(\mathtt{BODY}(pp) \cap \Ba))$

\hspace{2cm}$\mathtt{ACT}(\Ea)= \bigcup_{A \in \Ea} (\bigcup_{pp \in \mathtt{SUPPORT}(A)}(\mathtt{BODY}(pp) \cap \Aa))$

\hspace{2cm}$\mathtt{GOA}(\Ea)= \bigcup_{A \in \Ea} (\bigcup_{pp \in \mathtt{SUPPORT}(A)}(\mathtt{BODY}(pp) \cap \Ga))$

\hspace{2cm}$\mathtt{REC}(\Ea)= \bigcup_{A \in \Ea} (\bigcup_{pp \in \mathtt{SUPPORT}(A)}(\mathtt{BODY}(pp) \cap \Ra\Ea\Sa_{qua}))$\\

It is important to understand the following definition before presenting the results because both consistency and closure are specified based on justified conclusions. A goal that is the conclusion of an argument in any extension can be regarded as a justified conclusion, even if it is not in all extensions. From a more restrictive point of view, a goal can be regarded as a justified conclusion when it is the conclusion of an argument that belongs to all the extensions. 

\begin{defn}\textbf{(Justified conclusions)} Let $\Aa\Fa_g'=\langle Arg, \Ra_g \rangle$ be a general AF and after considering the successful attacks $\{\Ea_1, ..., \Ea_n\}$ ($n \geq 1$) be its set of extensions under the conflict-free semantics.

\begin{itemize}
\item $\mathtt{CONCS}(\Ea_i)=\{\mathtt{CLAIM}(A) | A \in \Ea_i\} (1 \leq i \leq n )$.
\item $\mathtt{Output}=\bigcap_{i=1,...,n} \mathtt{CONCS}(\Ea_i)$.
\end{itemize}

$\mathtt{CONCS}(\Ea_i)$ denotes the justified conclusions for a given extension $\Ea_i$ and $\mathtt{Output}$ denotes the conclusions that are supported by at least one argument
in each extension.
\end{defn}

An important property required in \cite{caminada2007evaluation} is direct consistency. An argumentation system satisfies direct consistency if its set of justified conclusions and the different sets of conclusions corresponding to each extension are consistent. This property is important in our approach because it guarantees that the agent will only pursue non-conflicting goals. 

\begin{theorem} \label{teorconst}\textbf{(Direct consistency)} Let $\Aa\Fa_g'=\langle \Aa rg, \Ra_g\rangle$ be a general AF after considering the successful attacks and $\Ea_1, ..., \Ea_n$ its conflict-free extensions. $\forall \Ea_i, i=1,...,n$, it holds that:

\begin{enumerate}
\item The set of beliefs $\mathtt{BEL}(\Ea_i)$ is a consistent set of literals.
\item The set of actions $\mathtt{ACT}(\Ea_i)$ is a consistent set of literals.
\item The set of goals $\mathtt{GOA}(\Ea_i)$ is a consistent set of literals.
\item The set of resources $\mathtt{REC}(\Ea)$ is a resource-consistent subset of $\Ra\Ea\Sa_{qua}$.
\item The set of goals $\mathtt{GOA}(\Ea_i)$ has no superfluous conflicting goals.
\end{enumerate}

\end{theorem}

Next property is closure, the idea of closure is that the set of justified conclusions of every extension should be closed under the set of plan rules $\Pa\Ra$. That is, if $g$ is a conclusion of an extension and there exists a plan rule $g \rightarrow g'$, then $g'$ should also be a conclusion of the same extension. Next definition states the closure of the set of plan rules of the agent.

\begin{defn}\label{defclos} (\textbf{Closure of $\Pa\Ra$)} Let $\Fa \subseteq \Ba \cup \Aa \cup \Ga \cup \Ra\Ea\Sa_{qua}$. The closure of $\Fa$ under the set $\Pa\Ra$ of plan rules, denoted by $\Ca l_{\Pa\Ra}(\Fa)$, is the smallest set such that:
\begin{itemize}
\item $\Fa \subseteq \Ca l_{\Pa\Ra}(\Fa)$
\item If $b_1, ..., b_n, g_1, ..., g_m, a_1, ..., a_l, res\_q(name,value)_1,..., res\_q(name,value)_k \rightarrow g \in \Fa$ and

 $b_1, ..., b_n, g_1, ..., g_m, a_1, ..., a_l, res\_q(name,value)_1,..., res\_q(name,value)_k \in \Ca l_{\Pa\Ra}(\Fa)$ then\break $g \in \Ca l_{\Pa\Ra}(\Fa)$.
\end{itemize}

If $\Fa = \Ca l_{\Pa\Ra}(\Fa)$, then $\Fa$ is said to be closed under the set $\Pa\Ra$.
\end{defn}

In our approach closure is important because it guarantees that all the goals that can be inferred from $\Pa\Ra$ be evaluated in terms of their possible conflicts, which in turn guarantees that the agent only will pursue non-conflicting goals.

\begin{theorem}\label{closure} {(Closure)} Let $\Pa\Ra$ be a set of plan rules, $\Aa\Fa_g'$ be an argumentation framework built from $\Pa\Ra$. $\mathtt{Output}$ is its set of justified conclusions, and $\Ea_1, ..., \Ea_n$ its extensions under the conflict-free semantics. $\Aa\Fa_g'$ satisfies closure iff:

\begin{enumerate} 
\item $\mathtt{CONCS}(\Ea_i)=\Ca l_{\Pa\Ra}(\mathtt{CONCS}(\Ea_i))$ for each $1 \leq i \leq n$.
\item $\mathtt{Output} = \Ca l_{\Pa\Ra}(\mathtt{Output})$.
\end{enumerate}

\end{theorem}

The last property our proposal should satisfy is indirect consistency. This property means that (i) the closure under the set of strict rules of the set of justified conclusions is consistent, and (ii) for each extension, the closure under the set of strict rules of its conclusions is consistent. 

\begin{theorem}\label{indirectconsit} \textbf{(Indirect consistency)} Let $\Pa\Ra$ be a set of plan rules, $\Aa\Fa_g'=\langle \Aa rg, \Ra_g \rangle$ be a general AF after considering the successful attacks. $\mathtt{OUTPUT}$ is its set of justified conclusions, and $\Ea_1, ..., \Ea_n$ its conflict-free extensions under the conflict-free semantics. $\Aa\Fa_g'=\langle \Aa rg, \Ra_g \rangle$ satisfies indirect consistency iff:

\begin{enumerate}
\item $\Ca l_{\Pa\Ra}(\mathtt{CONCS}(\Ea_i))$ is consistent for each $1 \leq i \leq n$.
\item $\Ca l_{\Pa\Ra}$ ($\mathtt{OUTPUT}$) is consistent.
\end{enumerate} 

\end{theorem}

\section{Related Work}
\label{correlatos}

We can divide the related works into two groups, the first one contains works whose approaches are not based on argumentation and the second one contains works that use argumentation to resolve the problem of goals selection. We begin presenting the first group of works.

One of the authors who has worked more on goals conflict -- especially on conflict detection -- is Thangarajah. In \cite{thangarajah2002representation}, he and his partners propose a general framework for conflict detection and resolution. For the conflict detection, they use a representation of the partial state of the world when each pursued goal is achieved, and by means of a mechanism they determine whether there are possible worlds where these the partial states co-exist. The conflict resolution is based on a preference ordering of the goals, which depends on both a priority-based commitment and an alternative-based commitment. In \cite{thangarajah2002avoiding}, they focus specifically on resource conflict. They characterize different types of resources and define resource requirements summaries. They also present mechanisms that allow agents to be aware of the possible and necessary resource requirements. This information is then used to detect and avoid conflicts. Ultimately, in \cite{thangarajah2003detecting}, they propose a mechanism that allows agents to detect and avoid a particular kind of conflict between goals, in which the achievement of a given goal undo conditions required for pursuing another goal. They use information about the requirements and resulting effects of goals and their associated plans to calculate and generate interaction summaries that let the agent schedule the execution of goals.

Other approaches worked on the selection strategy. Tinnemeier et al. \cite{tinnemeier2007goal} propose a mechanism to process incompatible goals by refraining the agent from adopting plans for goals that hinder the achievement of goals the agent is currently pursuing. They also define three types of goal selection strategies depending on the priority of the goals. Pokahr et al. \cite{pokahr2005goal} propose a goal deliberation strategy called Easy Deliberation, which allows agent programmers to specify the relationships between goals and enables an agent to deliberate about its goals by activating and deactivating certain goals. Their strategy enforces that only conflict-free goals are pursued and considers the order of the goals depending on their preference values. Wang et al. \cite{wang2012runtime} proposes a runtime goal conflict resolution model for agent systems, which consists of a goal state transition structure and a goal deliberation mechanism based on a extended event calculus.

Khan and Lespérance \cite{khan2010logical} present a framework where an agent can have multiple goals at different priority levels, possibly inconsistent with each other. The information about the goals priority is used to decide which of them should no longer be actively pursued in case they become mutually inconsistent. In their approach lower priority goals are not dropped permanently when they are inconsistent with other goals but they become inactive ones. In this way, when the world changes, the agent recomputes her adopted goals and some inactive goals may become active again. They argue that this ensures that an agent maximizes her utility

Finaly, Zatelli et al. \cite{zatelli2016conflicting} propose an approach to handle conflicting goals. They consider the conflicts in the plan level. In other words, they compare the plans that are used to achieve a given goal with the plans that are used to achieve another goal. They propose a technique 
that breaks down the plans in several sub-plans and specify the conflict between specific
sub-plans. Regarding the detection of conflicts, it is performed based on explicitly information about conflicting plans, which is introduced by the developer.

Next, we present the works that use argumentation techniques to resolve the problem of goals incompatibility.

In \cite{DBLP:journals/ijar/AmgoudDL11}, Amgoud presents a framework for handling contradictory desires. This framework is based on actions, which are defined as arguments. Thus, an action is made of a desire and a plan to achieve such desire. She defines four causes that could originate the conflicts among actions: (i) inconsistent desires, (ii) inconsistent plans, (iii) inconsistent consequences (it is referred to the consequences of performing an action), and (iv) inconsistency between a consequence and a plan. She uses trees for delineating the actions and identifying the attacks. Finally, she redefines the notion of defence and uses preferred extensions to return the different sets of desires that may be achieved together.

Hulstijn and van der Torre \cite{hulstijn2004combining} propose an approach that is mainly based on the framework of Amgoud \cite{DBLP:journals/ijar/AmgoudDL11} with some extensions. Like her, they use trees to represent the plans and work with the idea of conflicting actions in order to define the attack between plans and unlike her, they do not change the notion of defence but the notion of attack and exclude the possibility that their extensions contain two or more trees for the same desire.

The work of Atkinson and Bench-Capon \cite{atkinson2007action} deserves a special mention. The authors address the problem of practical reasoning based on presumptive argumentation using argument schemes and critical questions. Some of these critical questions have to do with to the action selection stage, specifically two of these questions are directly related to conflicts that may arise between actions. Such conflicts have similar characteristics than the terminal incompatibility and the superfluity formalized in this work.

Lastly, Amgoud et al. \cite{amgoud2008constrained} propose an argumentation system for practical reasoning. The proposed system is based on constrained argumentation \hbox{systems \cite{coste2006constrained}.} These systems extend the proposal of Dung \cite{dung1995acceptability} by adding constraints to the arguments, which need to be satisfied by the extensions returned by the system. One part of their work focuses on dealing with incompatibilities among plans, thus the possible conflicts arise when: (i) two plans cannot be executed at the same time, (ii) the execution of two plans will lead to contradictory states of the world, and (iii) the execution of a plan will prevent the execution of the second plan in the future. In order to deal with the conflicts they use e C-preferred semantics and C-stable \hbox{semantics \cite{coste2006constrained}.}

In this work, we focus basically on identifying conflicts between goals and on selecting the set of goals the agent will pursue. We consider three forms of incompatibilities based on the theory introduced in \cite{castelfranchi2007role}. Table \ref{tablacorr} shows a summary of all the related works along with their main characteristics; thus, column \textit{Topic} indicates the problem (or problems) that is (are) studied in the article, column \textit{Incompatibility} specifies the kind of conflict that is addressed. Notice that in most of the cases, this conflict is similar to the terminal incompatibility. The last column, \textit{Pref.}, indicates whether or not the preference (or priority, or importance) is taken into account for the selection of goals. 

\begin{table}[!htb]

\noindent\begin{tabular}{|c|c|c|c|}
\hline 
\textbf{Related work} & \textbf{Topic} & \textbf{Incompatibility} & \textbf{Pref.} \\ 
\hline 
Thangarajah et al. \cite{thangarajah2002representation} & detection & similar to terminal& yes \\ 
& selection & & • \\ 
\hline 
Thangarajah et al. \cite{thangarajah2002avoiding} & detection & resources & not  \\ 
& avoiding & & applies \\ 
\hline 
Thangarajah et al. \cite{thangarajah2003detecting} & detection & other & not  \\ 
& avoiding & & applies \\ 
\hline 
Tinnemeier et al. \cite{tinnemeier2007goal} & avoiding & similar to terminal & yes \\ 
\hline 
Pokahr et al. \cite{pokahr2005goal} & selection & not specified & yes \\ 
\hline 
Wang et al. \cite{wang2012runtime} & selection & not specified & not \\ 
\hline 
Khan and Lespérance \cite{khan2010logical} & selection & not specified & yes \\ 
\hline 
Zatelli et al. \cite{zatelli2016conflicting} & selection & similar to terminal & not \\ 
\hline 
\hline 
Amgoud \cite{DBLP:journals/ijar/AmgoudDL11} & detection & similar to terminal & not \\ 
& selection & & • \\ 
\hline 
Hulstijn and van der Torre \cite{hulstijn2004combining} & detection & similar to terminal & not \\ 
& selection &superfluity &  \\ 
\hline 
Amgoud et al. \cite{amgoud2008constrained} & detection & similar to terminal & not \\ 
& selection & & • \\ 
\hline 

Atkinson and Bench-Capon \cite{atkinson2007action} & detection & similar to terminal & not \\ 
&  &similar to superfluity & • \\ 
\hline 
Rahwan and Amgoud \cite{rahwan2006argumentation} & detection & similar to terminal & not \\ 
& selection & resources & • \\ 
\hline 
\end{tabular} 
\label{tablacorr}
\caption{Summary of the main related works along with the characteristics that will allow us to compare them with the approach proposed in this article. Argumentation-based works are separated from the others by using a double line. }

\end{table}
While it is true that most of the works focus on detecting and/or dealing with the problem of goals selection, we can notice that some other works focus on avoiding this problem, which is another way of addressing this problem. Avoiding a conflict is possible in some situation, for example, when the agent has a plan that is already being executed. In this case, the plans of new activated goals can be compared with the plan that is being executed. However, there could be other situations in which the agent needs to choose what goals he will pursue from a set of pursuble goals. We think that both avoiding and selecting may be complementary in order to deal with the incompatibility problem. Therefore, this is a good future direction of research.

Regarding the kinds of incompatibility the related works take into account. We notice that most of them consider a kind of conflict very similar to the terminal incompatibility, only two works deal with the resource conflicts, and two include a conflict similar to superfluity. With respect to the measure that allows defining an ordering of the goals, some works use the preference, some others the priority and others the importance value. We can notice that the works based on argumentation, unlike this work, do not use any of these measures. This has, in fact, an impact on the extensions returned by the semantics and it is a clear difference of our work.

Next we make more detailed analisys of the similarities and differences between our work and the work of Rahwan and Amgoud \mbox{\cite{rahwan2006argumentation}} since it is the closest related work to our approach.

  Although we use the same tree structure of the instrumental arguments, we use additional elements in the body of the plan rules. Thus, besides resources and goals, we also consider actions and beliefs.  As stated in Table 2, they define a set of conflicts that are similar to the notion of terminal incompatibility. For our part, we formalized a different form of incompatibility named superfluity, which has a different nature than any other type of conflict defined for arguments; however, in the domain of planning we consider that this is valid, as we explained in the corresponding section, and also defined a conflict based on the availability of resources of the agent. Thus, the difference lies in the way we use the knowledge about resources. In their case, they use it for determining a plan-plan conflict\footnote{Given two partial plans $[H, h]$ and $[H', h']$, a plan-plan conflict arises when $H \cup H' \vdash \perp$.}, which is directly related to logical inconsistency. In our case, the resource conflict has no relation with classical logical inconsistency, but with the amount of resources necessary for performing a plan and the availability of resources.

Regarding the strength calculation, we does not directly deal with measuring the strength of instrumental arguments; however, we use the preference value of the goals as a kind of strength value in order to define the defeat relation between arguments. 

So far, some similarities and differences between our work and the work in \cite{rahwan2006argumentation} has been mentioned. However, the main difference is related to the argumentative part of the work. Their proposal for obtaining the set of compatible goals is done at plans level and is based on the notion of conflict-freeness, maximality and utility. In our proposal, we also work at plans level and at goals level. At plans level, we also base on conflict-freeness, maximality and utility; however, our notion of maximality is different. While they take as acceptable a set of conflict-free extensions that is maximal for set inclusion, we take conflict-free extensions that maximize the number of achievable goals. We aim to either maximize the number of achievable goals or to maximize the utility of the agent in terms of achieving the most preferred goals, and the fact that a set of arguments is maximal for set inclusion does not necessarily imply that it will allow the agent to achieve as many goals as possible.

\section{Conclusions and Future Work}
\label{conclus}

This work presents the formalization and identification of three forms of incompatibility between procedural goals. We have noticed that the problem of selecting a set of compatible goals from a larger set of incompatible ones can be compared to the problem of calculating an extension in abstract argumentation. Therefore, we have adapted concepts of abstract argumentation to our
problem. In this adaptation, it was also considered the notion of defeat or successful attack since each goal has a different preference for the agent.

With respect to the research questions of the introduction, we can now state the following:

\begin{enumerate}
\item We have expressed a plan in terms of a instrumental argument, which is made up of a set of beliefs, actions, resources, and sub-goals in its premise, which have to be true, performed, available, and achieved, respectively, in order to the goal that is the claim of the argument. We defined different kinds of attacks according to the form of incompatibility we wanted to identify. Therefore, the answer to our first research question is positive. 
\item Both instrumental arguments and attacks are put together in argumentation frameworks. We defined one argumentation framework for each form of incompatibility and a general argumentation framework that involves all the all the kinds of attacks. Based on the general AF, we can identify when there is an attack between two goals.
\item We also generated a new argumentation framework whose elements are the pursuable goals along with their respective sub-goals. The attack relation is derived from the identification of attacks between arguments. Besides, the preference of each goal is taken into account to break the symmetry of the relation, if possible. 
\end{enumerate}

Next we list some possible future research directions:

\begin{itemize}

\item As mentioned in Section \ref{correlatos}, another way of dealing with incompatibility between goals is by avoiding it before a goal becomes pursuable. It would be interesting to study if the proposed approach can be used in this situation and how to integrate both forms of dealing with the problem.
\item We have defined attacks between instrumental arguments. We could extend our approach by including other kinds of attacks like the standard undercut and rebuttal. 
\item The preference value can be seen as a strength measure that determines what attack is successful; however, there are other criteria that may be considered. Hence, the identification of these criteria and the study of how to integrate them in order to determine the strength of an instrumental argument is another natural line of research.
\item The general AF can be seen as an expansion\footnote{More details about expansion theory can be found in \cite{baumann2012normal} and \cite{baumann2014role}.} of the terminal, resource and superfluity AFs. A further analysis is required to determine what kind of expansion it is and the properties it satisfies.

\item  According to van Riemsdijk et al. \hbox{\cite{van2005semantics}}, the fact of including a temporal component in the representation of goals could be a way of reducing inconsistency. More precisely, it would reduce what might appear to be an inconsistency without considering the temporal information. For instance, consider goal $be(fixed)$ and goal $clean(5,5)$, suppose that the agent has a order for pursuing goals that states that $be(fixed)$ has to be achieved before $clean(5,5)$, no matter the type of failure of the robot. In this case, these two goals cannot be seen as incompatible since the agent does not have to choose between them. Another example is when two goals have to be achieved at different times. For instance, the agent has the goal of going to the maintenance service on Saturday and has the goal of cleaning the room today. Although he is pursuing both goals, these are not incompatible since they have to be achieved at different times. We can notice that temporal information may in fact impact on the deliberation stage; therefore, it would be interesting to study how to include and use this type of information in our proposal.
\item For the identification of attacks between arguments we analize the whole tree of the argument; however, only some parts of it may be analized if we consider that some sub-goals were already achieved. In such case, we can denote these already achieved goals as beliefs. This will indeed be interesting, for example, in contexts where replanning is necessary. 
\end{itemize}

\section*{Acknowledgement}
The first author is founded by CAPES (Coordena\c{c}\~{a}o de Aperfei\c{c}oamento de Pessoal de N\'{i}vel Superior).

 \appendix

 \section{Proofs of Propositions and Theorems}
 \label{apen-proofs}

\noindent \textbf{Proposition \ref{propsimt}} If $(A,B) \in \Ra_t$, then $(B, A) \in \Ra_t$.

\begin{proof}

By reduction \textit{ab absurbo}. Assume that $(B,A) \notin \Ra_t$; hence, there is no partial plan $[H, \psi] \in \mathtt{SUPPORT}(B)$ and there is no partial plan\break $[H', \psi'] \in \mathtt{SUPPORT}(A)$ such that $\psi =  \neg '\psi$. Based on this, we can say that $\psi' =  \neg \psi$ does not hold either. However, if $\psi' = \neg \psi$ does not hold, it means that argument $A$ does not attack argument $B$ either, and hence $(A, B) \notin \Ra_t$, which contradicts the premise of the proposition.\hspace{4.5cm} $\Box$
 \end{proof}

\noindent \textbf{Proposition \ref{propsimr}} If $(A, B) \in \Ra_r$, then $(B, A) \in \Ra_r$.

\begin{proof} By reduction \textit{ab absurbo}. Suppose that $(B, A) \notin \Ra_r$; hence, there is no resource $res$ such that both argument $B$ and argument $A$ need it. Based on this, we can say that argument $A$ does not attack $B$ either; therefore,\break $(A, B) \notin \Ra_r$, which contradicts the premise of the proposition.\hspace{2.4cm}$\Box$
\end{proof}

\noindent \textbf{Proposition \ref{propsims}} If $(A, B) \in \Ra_s$, then $(B, A) \in \Ra_s$.

\begin{proof} We make a proof for each case:

\begin{itemize}
\item \textbf{Case 1:} By reduction \textit{ab absurbo}. Assume that $(B, A) \notin \Ra_s$, then $\mathtt{CLAIM}(A) \neq \mathtt{CLAIM}(B)$. This means that $(A, B) \notin \Ra_s$, which contradicts the premise of the proposition. 

\item \textbf{Case 2:} By reduction \textit{ab absurbo}. Assume that $(B, A) \notin \Ra_s$. Suppose that $\mathtt{CLAIM}(A) \neq \mathtt{CLAIM}(B)$, $\exists A',B' \in \Aa rg$ such that $[H'',\mathtt{CLAIM}(A)] \in \mathtt{SUPPORT}(A') $ and $[H''',\mathtt{CLAIM}(B)] \in \mathtt{SUPPORT}(B')$. Also suppose that $(B',A') \notin \Ra_s$. Due to first case, we can say that $(A',B') \notin \Ra_s$, which means that $(A, B) \notin \Ra_s$. This contradicts the premise of the proposition.

\item \textbf{Case 3:} By reduction \textit{ab absurbo}. Assume that $(B, A) \notin \Ra_s$. Suppose that $\mathtt{CLAIM}(A) \neq \mathtt{CLAIM}(B)$, $\exists A'\in \Aa rg$ such that $[H'',\mathtt{CLAIM}(A)] \in \mathtt{SUPPORT}(A')$. Also suppose that $(B,A') \notin \Ra_s$. Due to first case, we can say that $(A',B) \notin \Ra_s$, which means that $(A, B) \notin \Ra_s$. This contradicts the premise of the proposition. $\Box$
\end{itemize}

\end{proof}

\noindent \textbf{Proposition \ref{equivter}}
If $A \equiv_l A'$ and $B \equiv_l B'$ then $(A,B) \in \Ra_{x} =(A',B') \in \Ra_{x}$ (for $x \in \{t,s\}$).

\begin{proof} 
 
\textit{For partial-plans rebuttal} ($\Ra_t$). Consider that $(A,B) \in \Ra_{t}$, this means that there exists a partial plan $[H, \psi] \in \mathtt{SUPPORT}(A) $ and exists a partial plan $[H', \psi'] \in \mathtt{SUPPORT}(B)$ such that $\mathtt{HEAD}([H, \psi]) = \neg \mathtt{HEAD}([H', \psi'])$. Due to the definition of logical equivalence, we can say that given that $\mathtt{SUPPORT}(A)=\mathtt{SUPPORT}(A')$ then $[H, \psi] \in \mathtt{SUPPORT}(A') $ and given that\break $\mathtt{SUPPORT}(B)=\mathtt{SUPPORT}(B')$ then $[H', \psi'] \in \mathtt{SUPPORT}(B') $. Therefore, we can say that $(A',B') \in \Ra_t$.

\noindent\textit{For the superfluous attack} ($\Ra_s$). Let us present a proof for each case:

\begin{itemize}
\item \textbf{Case 1:} Since $A \equiv_l A'$ and $B \equiv_l B'$, when we state that $\mathtt{CLAIM}(A') = \mathtt{CLAIM}(B')$  is tantamount to saying that $\mathtt{CLAIM}(A)= \mathtt{CLAIM}(B)$ and when we state that $\mathtt{SUPPORT}(A') \neq \mathtt{SUPPORT}(B')$  is tantamount to saying that $\mathtt{SUPPORT}(A) \neq \mathtt{SUPPORT}(B)$. Therefore, we can say that $(A,B) \in \Ra_s = (A',B') \in \Ra_s$.

\item \textbf{Case 2:} Since $A \equiv_l A'$ and $B \equiv_l B'$ when we state that $\mathtt{CLAIM}(A') \neq \mathtt{CLAIM}(B')$ is tantamount to saying that $\mathtt{CLAIM}(A)\neq \mathtt{CLAIM}(B)$. When we state that $[H'',\mathtt{CLAIM}(A')] \in \mathtt{SUPPORT}(A'')$ is tantamount to saying that $[H'',\mathtt{CLAIM}(A)] \in \mathtt{SUPPORT}(A'')$ and when we state that $[H''',\break\mathtt{CLAIM}(B')] \in \mathtt{SUPPORT}(B'')$ is tantamount to saying that $[H''',\mathtt{CLAIM}(B)]\break \in \mathtt{SUPPORT}(B'')$ where $A'', B'' \in \Aa rg$ are two arguments such that\break $(A'',B'') \in \Ra_s$ due to the first case of superfluity. Therefore, we can say that $(A,B) \in \Ra_s = (A',B') \in \Ra_s$.

\item \textbf{Case 3:} Since $A \equiv_l A'$ and $B \equiv_l B'$ when we state that $\mathtt{CLAIM}(A') \neq \mathtt{CLAIM}(B')$ is tantamount to saying that $\mathtt{CLAIM}(A)\neq \mathtt{CLAIM}(B)$. When we state that $[H'',\mathtt{CLAIM}(A')] \in \mathtt{SUPPORT}(A'')$ is tantamount to saying that $[H'',\mathtt{CLAIM}(A)] \in \mathtt{SUPPORT}(A'')$ where $A'' \in \Aa rg$ is an argument such that $(A'',B) \in \Ra_s$ due to the first case of superfluity. Therefore, we can say that $(A,B) \in \Ra_s = (A',B') \in \Ra_s$.

\end{itemize}

Due to Proposition \ref{propsimt} and Proposition \ref{propsims}, these proofs also hold for $(B,A) \in \Ra_{x} =(B',A') \in \Ra_{x}$.$\Box$
\end{proof}

\noindent \textbf{Proposition \ref{equivres}} If $A \equiv_r A'$ and $B \equiv_r B'$ then $(A,B) \in \Ra_r=(A',B') \in \Ra_r$.

\begin{proof}

Let $\mathtt{REC}(A)$ be the resources necessary for performing the plan represented by argument $A$ and $\mathtt{REC}(B)$ be the resources necessary for performing the plan represented by argument $B$. Given that $A \equiv_r A'$ and $B \equiv_r B'$, it means that $\mathtt{REC}(A)=\mathtt{REC}(B)$. 

Since $(A,B) \in \Ra_r$, it means that $\exists res\_q(name,value) \in \mathtt{BODY}([H,\phi])$ --where $[H,\phi] \in \mathtt{SUPPORT}(A)$-- and $\exists res\_q(name,value)' \in \mathtt{BODY}([H',\phi'])$ --where $[H',\phi'] \in \mathtt{SUPPORT}(B)$-- such that $\Ra\Ea\Sa_{sum} \nvdash_r (res\_q(name,value) \wedge res\_q(name,value)')$. Considering that both resources $res\_q(name,value)$ and $res\_q(name,value)'$ are also part of the supports of arguments $A'$ and $B'$, we can say that $(A,B) \in \Ra_r=(A',B') \in \Ra_r$.

Due to Proposition \ref{propsimr}, this proof also hold for $(B,A) \in \Ra_r = (B',A') \in \Ra_r$.$\Box$
\end{proof}

\noindent \textbf{Proposition \ref{equivall}} If $A \equiv_w A'$ and $B \equiv_w B'$ then $(A,B) \in \Ra_{x}$=$(A',B') \in \Ra_{x}$ (for $x \in \{t,s,r\}$).

\begin{proof} This follows directly from the proofs of propositions \ref{equivter} and \ref{equivres}. \hspace{1.1cm}$\Box$

\end{proof}

\noindent \textbf{Proposition \ref{equivabres}} If $(A,B) \in \Ra_r $ and $A \equiv_r  A'$ then $(A',B) \in \Ra_r$.

\begin{proof}

Since $(A,B) \in \Ra_r$, it means that $\exists res\_q(name,value) \in \mathtt{BODY}([H,\phi])$ --where $[H,\phi] \in \mathtt{SUPPORT}(A)$-- and $\exists res\_q(name,value)' \in \mathtt{BODY}([H',\phi'])$ --where $[H',\phi'] \in \mathtt{SUPPORT}(B)$-- such that $\Ra\Ea\Sa_{sum} \nvdash_r (res\_q(name,value) \wedge res\_(name,value)')$. Consider that  $\exists (res\_q(name,value)_1 \in \mathtt{BODY}([H,\phi]')$ -- where $[H,\phi]' \in \mathtt{SUPPORT}(A')$.
Due to resource equivalence definition, we can say that $res\_q(name,value)_1=res\_q(name,value)$. Therefore, $(A',B) \in \Ra_r.$ \hspace{12.5cm} $\Box$
\end{proof}

\noindent \textbf{Proposition \ref{equivabx}} If $(A,B) \in \Ra_x$ and $A \equiv_l  A'$ then $(A',B) \in \Ra_x$ (for $x \in \{t,s\}$).

\begin{proof}\textit{For the partial-plans rebuttal }($\Ra_t$). By reduction \textit{ab absurbo}. Consider that there is no partial-plans rebuttal relation between $A'$ and $B$; hence, we can say that $\mathtt{SUPPORT}(A') \cup \mathtt{SUPPORT}(B) \nvdash \perp$. Since $A' \equiv A$, it means that $\mathtt{SUPPORT}(A')=\mathtt{SUPPORT}(A)$. Based on this, we can say that $\mathtt{SUPPORT}(A) \cup \mathtt{SUPPORT}(B) \nvdash \perp$, which means that there is no partial-plans rebuttal relation between $A$ and $B$. This contradicts the premise that there is an attack relation between arguments $A$ and $B$.\\

\noindent\textit{For the superfluous attack} ($\Ra_s$). Let us present a proof for each case. Firstly, let $A'', B'' \in \Aa rg$ be two arguments such that $(A'',B'') \in \Ra_s$ due to the first case of superfluity.

\begin{itemize}
\item \textbf{Case 1:} Since $(A,B) \in \Ra_s$ and $A \equiv_l A'$, we can say that $\mathtt{CLAIM}(A') = \mathtt{CLAIM}(B)$ and $\mathtt{SUPPORT}(A') \neq \mathtt{SUPPORT}(B)$. Therefore, we can say that $(A',B) \in \Ra_s$.

\item \textbf{Case 2:} Since $(A,B) \in \Ra_s$ and  $A \equiv_l A'$ we can say that $\mathtt{CLAIM}(A') \neq \mathtt{CLAIM}(B)$. Consider that $\exists A'', B'' \in \Aa rg$ such that $(A'',B'') \in \Ra_s$ due to the first case of superfluity. Also consider that $[H'',\mathtt{CLAIM}(A')] \in \mathtt{SUPPORT}(A'')$ and $[H'',\mathtt{CLAIM}(B)] \in \mathtt{SUPPORT}(B'')$. Therefore, we can say that $(A',B) \in \Ra_s$.

\item \textbf{Case 3:} Since $(A,B) \in \Ra_s$ and  $A \equiv_l A'$ we can say that $\mathtt{CLAIM}(A') \neq \mathtt{CLAIM}(B)$. Consider that $\exists A'' \in \Aa rg$ such that $(A'',B) \in \Ra_s$ due to the first case of superfluity. Also consider that that $[H'',\mathtt{CLAIM}(A')] \in \mathtt{SUPPORT}(A'')$ and $[H'',\mathtt{CLAIM}(A')] \notin \mathtt{SUPPORT}(B)$. Therefore, we can say that $(A',B) \in \Ra_s$.$\Box$

\end{itemize}

\end{proof}

\noindent \textbf{Proposition \ref{equivabw}} If $(A,B) \in \Ra_x$ and $A \equiv_w  A'$ then $(A',B) \in \Ra_x$, for $x \in \{t,r,s\}$.

\begin{proof} This follows directly from the proofs of propositions \ref{equivabres} and \ref{equivabx}. \hspace{1.1cm}$\Box$

\end{proof}

\noindent \textbf{Proposition \ref{propsinnum1}} If $(A,B) \in \Ra_t$, then  $\exists [H,\psi] \in \mathtt{SUPPORT}(A)$ and $\exists [H',\psi'] \in \mathtt{SUPPORT}(B)$ such that $\mathtt{HEAD}([H,\psi]) \cup \mathtt{HEAD}([H',\psi']) \vdash \perp$.

\begin{proof} This follows directly from the fact that in the attack defined for terminal incompatibility, i.e., partial-plans rebuttal, $\mathtt{HEAD}([H,\psi]) = \neg \mathtt{HEAD}([H',\psi'])$. Therefore, the union of both produces a contradiction. \hspace{0.4cm}$\Box$

\end{proof}

\noindent \textbf{Proposition \ref{propsinnum2}} Let $g,g' \in \Ga'$, $\mathtt{ARG}(g)$ and $\mathtt{ARG}(g')$ be the sets of arguments whose claims are $g$ and $g'$, respectively. If $\exists A \in \mathtt{ARG}(g)$ and $\exists B \in \mathtt{ARG}(g')$ such that $A$ and $B$ are conflict-free, then $(g,g'),(g',g) \not\in \Ra\Ga$.

\begin{proof} Suppose that all the arguments of $\mathtt{ARG}(g)$ and $\mathtt{ARG}(g')$ have a conflict relation except arguments $A$ and $B$, because they are conflict-free. This means that the conflict-free semantics returns an extension whose elements are arguments $A$ and $B$: $\Sa_{\Ca_\Fa}=\{\{A,B\}\}$. We then apply function $\mathtt{COMP\_GOALS}$ in order to obtain the set of compatible goals:\break $\mathtt{COMP\_GOALS}(\{A,B\})=\{g,g'\}$. Since $g$ and $g'$ are compatible, it means that there is no attack relation between them.

\end{proof}

\noindent \textbf{Theorem \ref{teorconst}} \textbf{(Direct consistency)} Let $\Aa\Fa_g'=\langle \Aa rg, \Ra_g\rangle$ be a general AF and $\Ea_1, ..., \Ea_n$ its conflict-free extensions. $\forall \Ea_i, i=1,...,n$, it holds that:

\begin{enumerate}
\item The set of beliefs $\mathtt{BEL}(\Ea_i)$ is a consistent set of literals.
\item The set of actions $\mathtt{ACT}(\Ea_i)$ is a consistent set of literals.
\item The set of goals $\mathtt{GOA}(\Ea_i)$ is a consistent set of literals.
\item The set of resources $\mathtt{REC}(\Ea)$ is a resource-consistent subset of $\Ra\Ea\Sa_{qua}$.
\item The set of goals $\mathtt{GOA}(\Ea_i)$ has no superfluous conflicting goals.
\end{enumerate}

\begin{proof} Let $\Ea_i$ be an extension of $\Aa\Fa_g'$.

\begin{enumerate}
\item Let us show that $\mathtt{BEL}(\Ea_i)$ is a consistent set of literals. 

Suppose that $\mathtt{BEL}(\Ea_i)$ is inconsistent. This means that $\exists\;b, \neg b \in \mathtt{BEL}(\Ea_i)$. Consider that $\exists\; A, B \in \Ea_i$ such that it happens that $[\{\},b] \in \mathtt{SUPPORT}(A)$ and $[\{\},\neg b] \in \mathtt{SUPPORT}(B)$. This means that there is a partial-plans rebuttal between $A$ and $B$ due to $b$ and $\neg b$. Since it contradicts the fact that $\Ea_i$ is conflict-free, we can say that $\mathtt{BEL}(\Ea_i)$ is consistent. 

\item Let us show that $\mathtt{ACT}(\Ea_i)$ is a consistent set of literals. 

Suppose that $\mathtt{ACT}(\Ea_i)$ is inconsistent. This means that $\exists\;a, \neg a \in \mathtt{BEL}(\Ea_i)$. Consider that $\exists\; A, B \in \Ea_i$ such that it happens that $[\{\},a] \in \mathtt{SUPPORT}(A)$ and $[\{\},\neg a] \in \mathtt{SUPPORT}(B)$. This means that there is a partial-plans rebuttal between $A$ and $B$ due to $a$ and $\neg a$. Since it contradicts the fact that $\Ea_i$ is conflict-free, we can say that $\mathtt{ACT}(\Ea_i)$ is consistent. 

\item Let us show that $\mathtt{GOA}(\Ea_i)$ is a consistent set of literals. 

Suppose that $\mathtt{GOA}(\Ea_i)$ is inconsistent. This means that $\exists\;g, \neg g \in \mathtt{SUB}(\Ea_i)$. Consider that $\exists\; A, B \in \Ea_i$ such that it happens that $[\{\},g] \in \mathtt{SUPPORT}(A)$ and $[\{\},\neg g] \in \mathtt{SUPPORT}(B)$. This means that there is a partial-plans rebuttal between $A$ and $B$ due to $g$ and $\neg g$. Since it contradicts the fact that $\Ea_i$ is conflict-free, we can say that $\mathtt{GOA}(\Ea_i)$ is consistent. 

\item Let us show that $\mathtt{REC}(\Ea_i)$ is a resource-consistent subset of $\Ra\Ea\Sa_{qua}$. 

Suppose that $\mathtt{REC}(\Ea_i)$ is resource-inconsistent. This means that $\exists\;res\_q(name,value)_A, res\_q(name,value)_B \in \mathtt{REC}(\Ea_i)$ such that $\Ra\Ea\Sa_{sum} \nvdash_r\; (res\_q(name,value)_A \wedge res\_q(name,value)_B)$. This in turn means that $\exists\; A, B \in \Ea_i$ such that there is a resource attack between $A$ and $B$. This contradicts the fact that $\Ea_i$ is conflict-free and therefore, we can say that $\mathtt{REC}(\Ea_i)$ is a resource-consistent.

\item Let us show that $\mathtt{GOA}(\Ea_i)$ has no superfluous conflicting goals. 

Suppose that $\exists g, g' \in \mathtt{GOA}(\Ea_i)$ such that $g$ and $g'$ are superfluous goals. Let us analize each case of the superfluous attack:

\begin{itemize}
\item \textbf{Case 1:} Since $g$ and $g'$ are superfluous goals, it means that $\exists A, B \in \Ea_i$ such that $\mathtt{CLAIM}(A)=\mathtt{CLAIM}(B)=g=g'$ and $\mathtt{SUPPORT}(A)\neq\mathtt{SUPPORT}(B)$. Hence, there is a superfluous attack between $A$ and $B$. This contradicts the fact that $\Ea_i$ is conflict free.

\item \textbf{Case 2:} Given that $g, g' \in \mathtt{GOA}(\Ea_i)$, it means that $\exists C,D \in \Ea_i$ such that $\mathtt{CLAIM}(C)=g$ and  $\mathtt{CLAIM}(D)=g'$. Consider now that $\exists g'' \in \mathtt{GOA}(\Ea_i)$ and $\exists A \in \Ea_i$ such that $\mathtt{CLAIM}(A)=g''$. Also consider that $\exists B \in \Aa rg$ such that $B \notin \Ea_i$ because $\mathtt{CLAIM}(B) = \mathtt{CLAIM}(A)$ and $\mathtt{SUPPORT}(A)\neq\mathtt{SUPPORT}(B)$, which means that there is a superfluous attack between $A$ and $B$. Since $g$ and $g'$ are superfluous goals, it means that $g \neq g'$, $g \in \mathtt{BODY}(\mathtt{SUPPORT}(A))$ and $g' \in \mathtt{BODY}(\mathtt{SUPPORT}(B))$. Thus, we can say that there is a superfluous attacks between $C$ and $D$, which contradicts the fact that $\Ea_i$ is conflict free.

\item \textbf{Case 3:} Given that $g, g' \in \mathtt{GOA}(\Ea_i)$, it means that $\exists C,B \in \Ea_i$ such that $\mathtt{CLAIM}(C)=g$ and  $\mathtt{CLAIM}(B)=g'$. Consider now that $\exists A \in \Aa rg$ such that $A \notin \Ea_i$ because $\mathtt{CLAIM}(A) = \mathtt{CLAIM}(B)$ and $\mathtt{SUPPORT}(A)\neq\mathtt{SUPPORT}(B)$, which means that there is a superfluous attack between $A$ and $B$. Since $g$ and $g'$ are superfluous goals, it means that $g \neq g'$, $g \in \mathtt{BODY}(\mathtt{SUPPORT}(A))$ and $g \notin \mathtt{BODY}(\mathtt{SUPPORT}(B))$. Thus, we can say that there is a superfluous attacks between $C$ and $B$, which contradicts the fact that $\Ea_i$ is conflict free.

\end{itemize}

\end{enumerate} 

Finally, since all the extensions obtained from $\Aa\Fa$ are consistent, then $\mathtt{OUTPUT}$ is also consistent.  $\Box$
\end{proof}

\noindent \textbf{Theorem \ref{closure} (Closure)} Let $\Pa\Ra$ be a set of plan rules, $\Aa\Fa_g'$ be an argumentation framework built from $\Pa\Ra$. $\mathtt{Output}$ is its set of justified conclusions, and $\Ea_1, ..., \Ea_n$ its extensions under the conflict-free semantics. $\Aa\Fa_g'$ satisfies closure iff:

\begin{enumerate} 
\item $\mathtt{CONCS}(\Ea_i)=\Ca l_{\Pa\Ra}(\mathtt{CONCS}(\Ea_i))$ for each $1 \leq i \leq n$.
\item $\mathtt{Output} = \Ca l_{\Pa\Ra}(\mathtt{Output})$.
\end{enumerate}

\begin{proof}

Let us call $\Aa rgCl$ the arguments that can be built from $\mathtt{BEL}(\Ea) \cup \mathtt{ACT}(\Ea) \cup \mathtt{GOA}(\Ea) \cup \mathtt{REC}(\Ea) $. 

Given that $\mathtt{CONCS}(\Ea_i)=\Ca l_{\Pa\Ra}(\mathtt{CONCS}(\Ea_i))$, we will proof that $\forall \Ea_i, i=1, ..., n$, it holds that $\Ea_i = \Aa rgCl$; thus, we will first proof that $\Ea_i \subseteq \Aa rgCl$ and then that $\Aa rgCl \subseteq \Ea_i$.

\begin{enumerate}
\item $\Ea_i \subseteq \Aa rgCl$: This is trivial.
\item $\Aa rgCl \subseteq \Ea_i$: Suppose that $\Aa rgCl \nsubseteq \Ea_i$; hence, $\exists A \in \Aa rgCl$ such that $A \not\in \Ea_i $. Since  $A \not\in \Ea_i $, it means that $\exists B \in \Ea_i$ such that $(B,A) \in \Ra_g$ or $(A,B) \in \Ra_g$. There are three situation in which this happens, one for each form of incompatibility:

\begin{enumerate}
\item \textit{When the attack is a partial-plans rebuttal} ($(B,A) \in \Ra_t$ or $(A,B) \in \Ra_t$): This means that $\exists [H, \psi] \in \mathtt{SUPPORT}(B)$ and $\exists [H', \psi'] \in \mathtt{SUPPORT}(A)$ such that $\psi = \neg \psi'$, considering that $\psi, \psi' \in \Ba$ or $\psi, \psi' \in \Ga$ or $\psi, \psi' \in \Aa$. Since both $A$ and $B$ are built from $\mathtt{BEL}(\Ea_i)$, $\mathtt{ACT}(\Ea_i)$, and $\mathtt{GOA}(\Ea_i)$ this would mean that $\mathtt{BEL}(\Ea_i)$, $\mathtt{ACT}(\Ea_i)$, or $\mathtt{GOA}(\Ea_i)$ are inconsistent, which contradicts the first, second, and third items of Theorem \ref{teorconst}.

\item \textit{When it is a resource attack} ($(B,A) \in \Ra_r$ or $(A,B) \in \Ra_r$): Let $res\_q(name,value)_A, res\_q(name,value)_B \in \Ra\Ea\Sa_{qua}$. Consider that $\exists [H, \psi] \in \mathtt{SUPPORT}(B)$ and $\exists [H', \psi'] \in \mathtt{SUPPORT}(A)$. Also consider that $\exists res\_q(name,value)_B \in \mathtt{BODY}([H, \psi])$ and $\exists res\_q(name,value)_A \in \mathtt{SUPPORT}([H', \psi'])$ such that $\Ra\Ea\Sa_{sum} \nvdash_r\;(res\_q(name,value)_B \wedge res\_q(name,value)_A)$. Therefore, there is a resource attack between $A$ and $B$. Since the resources for building $A$ and $B$ are part of $\mathtt{REC}(\Ea_i)$, this would mean that $\mathtt{REC}(\Ea_i)$ is resource-inconsistent, which contradicts the forth item of Theorem \ref{teorconst}. 

\item \textit{When it is a superfluous attack} ($(B,A) \in \Ra_s$ or $(A,B) \in \Ra_s$): Let $g=\mathtt{CLAIM}(A)$ and $g'=\mathtt{CLAIM}(B)$. Since there is a superfluous attack between $A$ and $B$, it means that $g$ and $g'$ are superfluous goals. Given that both arguments are built from $\mathtt{BEL}(\Ea_i)$, $\mathtt{ACT}(\Ea_i)$, $\mathtt{REC}(\Ea_i)$, and $\mathtt{GOA}(\Ea_i)$, we can say that their conclusions are also part of $\mathtt{GOA}(\Ea_i)$, i.e. $g,g' \in \mathtt{GOA}(\Ea_i)$. This contradicts the last item of Theorem \ref{teorconst}, which proofs that there is no superfluous conflicting goals in $\mathtt{GOA}(\Ea_i)$.$\Box$

\end{enumerate}

\end{enumerate}

\end{proof}

\noindent \textbf{Theorem \ref{indirectconsit}} \textbf{(Indirect consistency)} Let $\Pa\Ra$ be a set of plan rules, $\Aa\Fa_g'=\langle \Aa rg, \Ra_g \rangle$ be a general AF. $\mathtt{OUTPUT}$ is its set of justified conclusions, and $\Ea_1, ..., \Ea_n$ its conflict-free extensions under the conflict-free semantics. $\Aa\Fa_g'=\langle \Aa rg, \Ra_g \rangle$ satisfies indirect consistency iff:

\begin{enumerate}
\item $\Ca l_{\Pa\Ra}(\mathtt{CONCS}(\Ea_i))$ is consistent for each $1 \leq i \leq n$.
\item $\Ca l_{\Pa\Ra}$ ($\mathtt{OUTPUT}$) is consistent.
\end{enumerate}

\begin{proof}

Based on Proposition 7 defined in \cite{caminada2007evaluation}\footnote{Extracted from \cite{caminada2007evaluation}: ``\textbf{Proposition 7}. Let $\langle A,Def \rangle$ be an argumentation system. If $\langle A,Def \rangle$ satisfies closure and direct consistency, then it also satisfies indirect consistency.''.}, we can say that our proposal satisfies indirect inconsistency since it satisfies closure and direct consistency.

\end{proof}

\bibliographystyle{unsrt}  
\bibliography{sbc-template}  


\end{document}